\pdfoutput=1

\documentclass[11pt]{article}

\usepackage{EACL2023}

\usepackage{times}
\usepackage{latexsym}
\usepackage{booktabs}
\usepackage{graphics}
\usepackage{graphicx}
\usepackage{multirow}
\usepackage{array}
\usepackage{geometry}
\usepackage{xcolor}
\usepackage{float}

\usepackage[T1]{fontenc}

\usepackage[utf8]{inputenc}

\usepackage{microtype}

\usepackage{inconsolata}

\title{On Evaluation of Document Classification using RVL-CDIP}

\author{~~~~~~~~~~~Stefan Larson$^{1}$\thanks{~~~Corresponding email: \url{stefan.dataset@gmail.com}} \And
  Gordon Lim$^2$~\\\\
    \begin{tabular}{ccc}
       $^1$Vanderbilt University & & $^2$University of Michigan \\
      Nashville, TN, USA & & Ann Arbor, MI, USA
  \end{tabular}
  \And
  Kevin Leach$^1$~~~~~~~~~~
  }

\begin{document}
\maketitle
\begin{abstract}
The RVL-CDIP benchmark is widely used for measuring performance on the task of document classification.
Despite its widespread use, we reveal several undesirable characteristics of the RVL-CDIP benchmark.
These include (1) substantial amounts of label noise, which we estimate to be 8.1\% (ranging between 1.6\% to 16.9\% per document category); (2) presence of many ambiguous or multi-label documents; (3) a large overlap between test and train splits, which can inflate model performance metrics; and (4) presence of sensitive personally-identifiable information like US Social Security numbers (SSNs).
We argue that there is a risk in using RVL-CDIP for benchmarking document classifiers, as its limited scope, presence of errors (state-of-the-art models now achieve accuracy error rates that are within our estimated label error rate), and lack of diversity make it less than ideal for benchmarking.
We further advocate for the creation of a new document classification benchmark, and provide recommendations for what characteristics such a resource should include.
\end{abstract}

\section{Introduction}

\begin{figure}
    \centering\scalebox{0.51}{ 
    \includegraphics{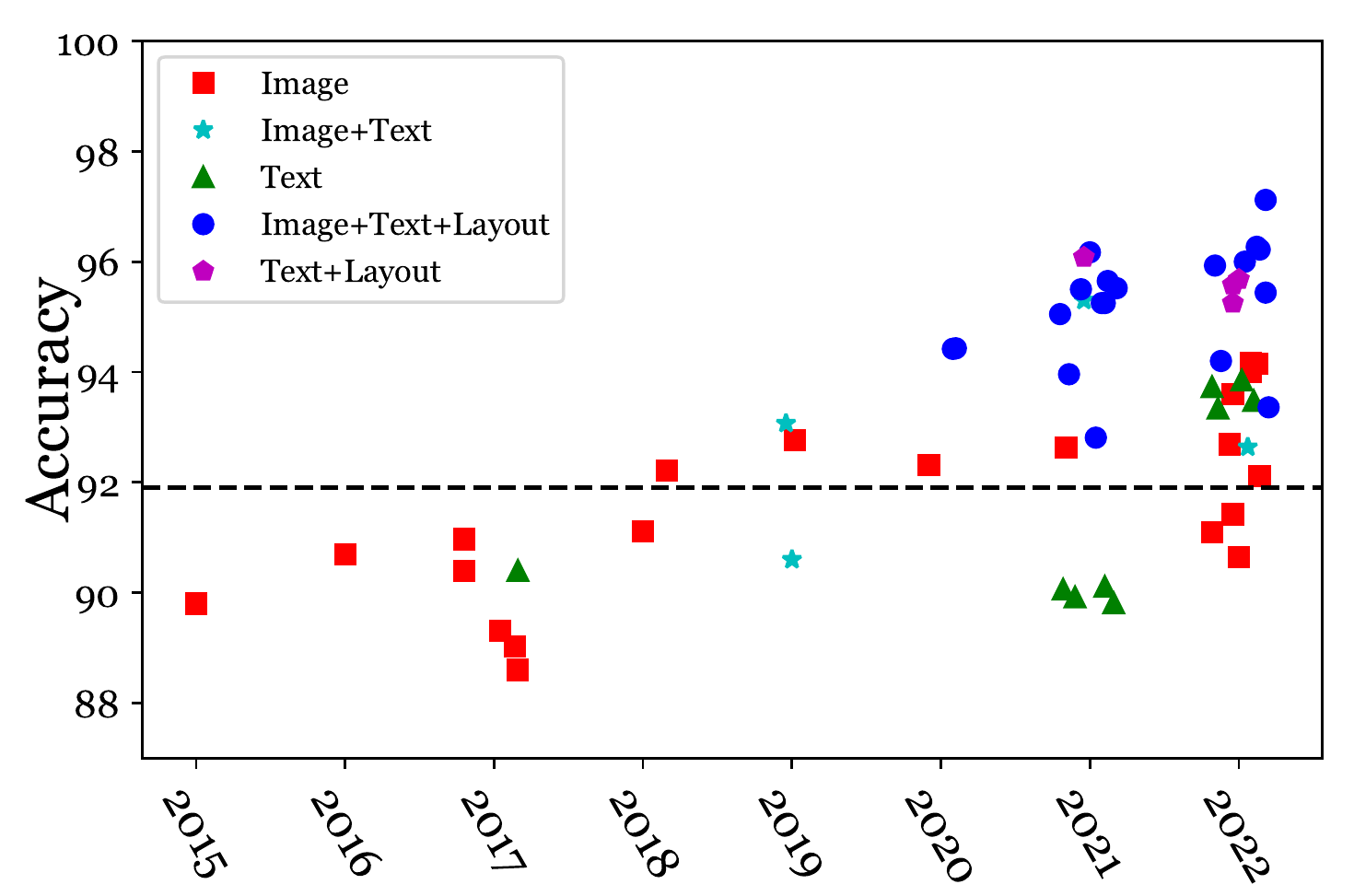}}
    \caption{Model accuracy on RVL-CDIP by year and modality. The horizontal dashed  line represents our estimated label error rate for RVL-CDIP's test set.}
    \label{fig:accuracy-by-year}
\end{figure}

Within the document understanding research area, the RVL-CDIP dataset \cite{harley2015icdar} has emerged as the primary benchmark for evaluating and comparing document classifiers.
RVL-CDIP is composed of 16 document type categories, including \texttt{resume}, \texttt{letter}, \texttt{invoice}, etc. Its large volume of training data---320,000 samples---facilitates benchmarking state-of-the-art deep learning and transformer-based architectures.
While initially released as a computer vision benchmark in 2015, more recent state-of-the-art models now incorporate image, text, and page layout modalities.
For instance, recent tri-modal models like DocFormer \cite{Appalaraju_2021_ICCV-docformer}, ERNIE-Layout \cite{ernie-layout-2022}, LayoutLMv3 \cite{huang2022layoutlmv3}, and Bi-VLDoc \cite{bi-vldoc-2021} now achieve classification accuracies ranging in the mid- to high-90s, with Bi-VLDoc reporting a state-of-the-art of 97.12\% on the RVL-CDIP test set.
This is a large improvement over earlier image-centric work, and we chart this improvement in Figure~\ref{fig:accuracy-by-year}.

As model performance on RVL-CDIP improves, it becomes increasingly important to ensure that further gains are meaningful with respect to the classification task.
This concern has been raised by prior work that has found that benchmark evaluation datasets often contain substantial amounts of label errors or noise \citep[e.g.,][]{test-errors-2021}, substantial overlap between test and train data \cite[e.g.,][]{elangovan-etal-2021-memorization, sogaard-etal-2021-need}, and data collection artifacts that cause models to overfit to spurious cues \citep[e.g.,][]{gururangan-etal-2018-annotation, mccoy-etal-2019-right}.
Therefore, we cast a critical eye to the RVL-CDIP benchmark to answer: \emph{Is RVL-CDIP still suitable for effectively measuring the performance of document classifiers?}

In doing so, we first observe a lack of clear label or annotation guidelines provided with the original introduction of RVL-CDIP.
Therefore, we create verifiable label guidelines for the 16 RVL-CDIP categories.
With these guidelines, we are then able to conduct a review of the data, and we find that label errors account for an estimated 8.1\% of data RVL-CDIP's test split, a rate greater than the current state-of-the-art model accuracy error rate, indicating that contemporary high-performing models are overfitting to noise.
We also observe relatively high rates of documents that have ambiguous or multiple valid labels, which is problematic given RVL-CDIP is a single-label classification benchmark.
Additionally, we also observe a large overlap between test and train data splits, where there are (near-) duplicate documents seen in both train and test splits, as well as documents that share common templates.
Lastly, our review of RVL-CDIP data uncovered a surprisingly large amount of sensitive personally-identifiable information, particularly in the \texttt{resume} category, where we found 7.7\% of documents contained US Social Security numbers.

We argue that the characteristics that we observe make RVL-CDIP an unattractive benchmark for training and evaluating document classifiers. We end with recommendations for what qualities a new document classification benchmark should have.

  \begin{table}[t!]
    \centering\scalebox{0.58}{ 
    \begin{tabular}{llc}
    \toprule
        \textbf{Model (Reported by)} & \textbf{Modality} & \textbf{Accuracy} \\
    \midrule
    Bi-VLDoc \cite{bi-vldoc-2021} & I, T, L& 97.12 \\
    ERNIE-Layout-large \cite{ernie-layout-2022} & I, T, L & 96.27\\
        UDOP-Dual \cite{udop-tang-unifying-2022} & I, T, L & 96.22\\
        DocFormer-base \cite{Appalaraju_2021_ICCV-docformer}  & I, T, L & 96.17\\
        StructuralLM-large \cite{li-etal-2021-structurallm} & T, L & 96.08 \\
        UDOP \cite{udop-tang-unifying-2022} & I, T, L & 96.00 \\
        LayoutLMv3-large \cite{huang2022layoutlmv3} & I, T, L & 95.93 \\
        LiLT-base \cite{wang-etal-2022-lilt} & T, L & 95.68\\
        LayoutLMv2-large \cite{xu-etal-2021-layoutlmv2} & I, T, L & 95.65 \\
        BROS-base \cite{wang-etal-2022-lilt} & T, L & 95.58 \\
        TILT-large \cite{TILT-2021} & I, T, L & 95.52 \\
        DocFormer-large \cite{Appalaraju_2021_ICCV-docformer} & I, T, L & 95.50\\
        LayoutLMv3-base \cite{huang2022layoutlmv3} & I, T, L & 95.44 \\
        Donut \cite{kim2021donut} & I, T & 95.30 \\
        Wukong-Reader-large \cite{wukong-reader-2022} & I, T, L & 95.26\\
        \citet{pham-2022-long} & T, L & 95.25 \\
        TILT-base \cite{TILT-2021} & I, T, L & 95.25\\
        LayoutLMv2-base \cite{xu-etal-2021-layoutlmv2} & I, T, L & 95.25 \\
        UDoc-star \cite{udoc-2021} & I, T, L & 95.05 \\
        Wukong-Reader-base \cite{wukong-reader-2022} & I, T, L & 94.91\\
        StrucTexTv2-large \cite{yu2023structextv} & I & 94.62 \\
        LayoutLMv1-base \cite{layoutlm-v1-2020-xu} & I, T, L & 94.43 \\
        LayoutLMv1-large \cite{layoutlm-v1-2020-xu} & I, T, L & 94.42 \\
        MATrIX \cite{matrix-2022} & I, T, L & 94.20 \\
        DocXClassifier-xl \cite{docxclassifier-saifullah2022} & I & 94.17 \\
        DocXClassifier-large \cite{docxclassifier-saifullah2022} & I & 94.15\\
        DocXClassifier-base \cite{docxclassifier-saifullah2022} & I & 94.00\\
        UDoc \cite{udoc-2021} & I, T, L & 93.96 \\
        Longformer-base \cite{pham-2022-long} & T & 93.85 \\
        Longformer-large \cite{pham-2022-long} & T & 93.73 \\
        MGDoc \cite{mgdoc-2022-emnlp} & I, T, L & 93.64 \\
        Dessurt \cite{dessurt-2022} & I & 93.60 \\
        Bigbird-base \cite{pham-2022-long} & T & 93.48 \\
        \citet{pramanik-2022} & I, T, L & 93.36 \\
        Bigbird-large \cite{pham-2022-long} & T & 93.34 \\
        Multimodal Ensemble \cite{dauphinee-2019} & I, T & 93.07 \\
        SelfDoc \cite{selfdoc-2021} & I, T, L & 92.81 \\
        LadderNet \cite{laddernet-2019} & I & 92.77 \\
        \citet{Zingaro2021} & I, T & 92.70 \\
        DiT-large \cite{li2022dit} & I & 92.69 \\
        VLCDoC \cite{vlcdoc} & I, T & 92.64 \\
        InceptionResNetV2 \cite{xu-etal-2021-layoutlmv2} & I & 92.63 \\
        EfficientNet \cite{ferrando-2020} & I & 92.31 \\
        Region Ensemble \cite{das-2018} & I & 92.21 \\
        DiT-base \cite{li2022dit} & I & 92.11 \\
        MAE-base \cite{li2022dit} & I & 91.42\\
        Stacked CNN Single \cite{das-2018} & I & 91.11\\
        BEiT-base \cite{li2022dit} & I & 91.09 \\
        VGG-16 \cite{afzal-2017} & I & 90.97 \\
        \citet{csurka-2016} & I & 90.70 \\
        ResNext-101 \cite{li2022dit} & I & 90.65 \\
        \citet{audebert-2019} & I, T & 90.60 \\
        ResNet-50 \cite{afzal-2017} & I & 90.40 \\
        RoBERTa-large \cite{li-etal-2021-structurallm} & T & 90.11\\
        RoBERTa-base \cite{li-etal-2021-structurallm} & T & 90.06\\
        BERT-large \cite{li-etal-2021-structurallm} & T & 89.92 \\
        BERT-base \cite{li-etal-2021-structurallm} & T & 89.81 \\
        \citet{tensmeyer-2017} & I & 89.31 \\
        GoogLeNet \cite{afzal-2017} & I & 89.02 \\
        AlexNet \cite{afzal-2017} & I & 88.60 \\
    \bottomrule
    \end{tabular}}
    \caption{Model accuracy on RVL-CDIP for various image (I), text (T), and layout-based (L) document classification models, ordered by reported score. Models incorporating multiple modalities typically outperform uni-modal models.}
    \label{tab:model-scores}
\end{table}

\section{Related Work}
This section discusses related work in two areas: (1) prior work in document classification on RVL-CDIP, and (2) prior work on analyzing datasets.

\subsection{RVL-CDIP and Document Classification}
The RVL-CDIP corpus has been used as a benchmark for document classification since its introduction by \citet{harley2015icdar}, who used it to evaluate convolutional neural network (CNN) image classifiers on the dataset's document images.
Most immediate follow-up work followed \citet{harley2015icdar} and explored different image-based CNN models, as done in \citet{csurka-2016,afzal-2017, tensmeyer-2017,das-2018, ferrando-2020}.
Just relying on image features is limited, as much of a document's "essence" is informed by its textual content.
Therefore, more recent work has incorporated the textual modality, including \citet{audebert-2019} and \citet{dauphinee-2019}.

\begin{figure*}
    \centering\scalebox{0.45}{
    \includegraphics{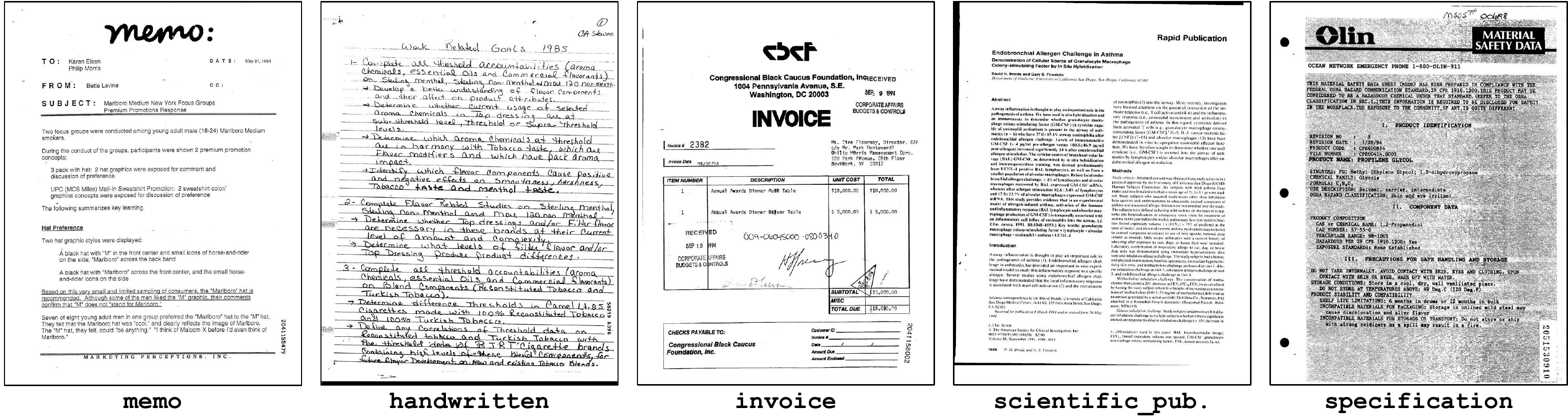}}
    \caption{Samples from the RVL-CDIP dataset.}
    \label{fig:rvlcdip_samples}
\end{figure*}

\begin{table}[]
     \centering\scalebox{0.65}{
     \begin{tabular}{cc}
     \toprule
      \texttt{advertisement} & \texttt{memo}\\
      \texttt{budget} &  \texttt{news\_article}\\
      \texttt{email} &  \texttt{presentation}\\
      \texttt{file\_folder} &  \texttt{questionnaire}\\
      \texttt{form} & \texttt{resume}\\
      \texttt{handwritten} & \texttt{scientific\_publication}\\
      \texttt{invoice} & \texttt{scientific\_report}\\
      \texttt{letter} & \texttt{specification}\\
      \bottomrule
     \end{tabular}}
     \caption{RVL-CDIP document type categories.}
     \label{tab:rvl-cdip-categories}
 \end{table}

Even more recent work has capitalized on the transformer model architecture, often combining vision transformers with large transformer-based language models (and often combining these with a third modality based on page layout derived from detected optical character recognition (OCR) regions) as in LayoutLMv1 \cite{layoutlm-v1-2020-xu}, LayoutLMv2 \cite{xu-etal-2021-layoutlmv2}, LayoutLMv3 \cite{huang2022layoutlmv3}, DocFormer \cite{Appalaraju_2021_ICCV-docformer}, TILT \cite{TILT-2021}, and ERNIE-layout \cite{ernie-layout-2022}.
These more recent transformer-based models have achieved state-of-the-art accuracy scores on RVL-CDIP, the most recent being \citet{bi-vldoc-2021}'s Bi-VLDoc, which achieves a reported accuracy of 97.12\% on RVL-CDIP.
(For a listing of models benchmarked on RVL-CDIP since \citet{harley2015icdar}, see Table~\ref{tab:model-scores}.)

Despite these high scores, recent work has exposed gaps in models trained on RVL-CDIP.
In particular, recent work has found that models trained on RVL-CDIP perform poorly on out-of-distribution data \cite{larson-2022-rvlcdip-ood} and perturbed in-distribution data \cite{saifullah-2022}. 
We also mention that RVL-CDIP is often used as a pre-training dataset, where models are first pre-trained on RVL-CDIP (and perhaps others) and then evaluated on other downstream tasks or datasets (e.g., \citet{nguyen-etal-2021-skim-attention, emmdocclassifier-2022}).
Other datasets like FUNSD are subsets of RVL-CDIP \cite{funsd}.

 

\subsection{Analysis of Datasets}
 
Prior work has investigated the presence of label and annotation errors and corpus quality in NLP and image datasets.
This work includes \citet{vldb-detecting-data-errors-2016, revisiting-oxford-and-paris-2018, muller-ijcnn-mislabeled-2019, mislabeled-pleiss-2020, test-errors-2021, kreutzer-etal-2022-quality, ying-thomas-2022-label-banking77, chong-etal-2022-detecting}.
One common conclusion is that the utility of a benchmark evaluation dataset is lessened if the label error and/or ambiguity rate is close to- or exceeds model prediction error rate.
This has been observed for various datasets, such as ATIS \cite{bechet-atis-shallow-2018, niu-penn-2019-rationally-atis}, and the CNN/Daily Mail reading comprehension task \cite{chen-etal-2016-thorough}.

Orthogonal to label errors, prior work has also observed non-trivial overlap between test and train splits in datasets on which natural language processing and computer vision models are evaluated \citep[e.g.,][]{finegan-dollak-etal-2018-improving, allamanis-2019-adverse-duplication, cifar-duplicates-2020,  lewis-etal-2021-question, wen-etal-2022-empirical, croft-2023-svp-data-quality}.
Such work often argues that non-trivial amounts of overlap between test and train data can lead to "inflated" performance scores, as overlapping data can reward a model's ability to memorize training data \cite{elangovan-etal-2021-memorization}, and to under-estimate out-of-sample error \cite{sogaard-etal-2021-need}.
Evidence of this can also be found in the multitude of studies that report lower model performance scores on newly-collected evaluation sets versus reported scores on benchmarks \citep[e.g.,][]{AUGENSTEIN201761, pmlr-v97-recht19a, harrigian-etal-2020-models, kim2022your, larson-2022-rvlcdip-ood}.
In this paper, we investigate the presence of errors, ambiguous data, and overlapping test-train data for the RVL-CDIP benchmark dataset.

\begin{table*}[t]
\centering
\scalebox{0.72}{
    \begin{tabular}{l m{18cm} }
        \toprule
        \textbf{Category} & \textbf{Description}\\
        \midrule
        \texttt{advertisement} & Advertisements from print-form media like newspapers and magazines. Also a small amount of scripts for television or radio advertisements. A small amount of "ad order instructions" and "ad insertion" documents. \\
        \midrule
        \texttt{budget} & Includes various budget documents such as expense, spending, sales, cash, and accounting reports and forecasts; budgets; quotes and estimates; and income and bank statements. Also includes receipt-like documents such as political campaign contribution requests and other receipts, as well as checks and check stubs. \\
        \midrule
        \texttt{email} & Scanned images of printed emails.  \\
        \midrule
        \texttt{file\_folder} & Scanned images of folders and binders. Folder scans are often characterized by vertically oriented text (indicating a folder label). A moderate amount of file folders in RVL-CDIP contain handwritten text or notes. Some scanned folders may be indistinguishable from blank pages. \\
        \midrule
        \texttt{form} & Form documents with form-like elements (e.g., lines or spaces for user-provided data entry). The form-like elements can appear empty or filled. \\
        \midrule
        \texttt{handwritten} & Includes handwritten documents like handwritten letters and scientific notes. \\
        \midrule
        \texttt{invoice} & Includes invoices, bills, and account statements. \\
        \midrule
        \texttt{letter} & Letters, often with letterhead and commonly with "Dear..." salutations. The distinction between letters and memos is often unclear in RVL-CDIP. \\
        \midrule
        \texttt{memo} & Memoranda or inter-office correspondence documents, often with clear "TO", "FROM", "SUBJECT" headings. \\
        \midrule
        \texttt{news\_article} & Includes news articles in the form of clippings from newspapers and other print-form news media, as well as a small amount of news articles from the web. \\
        \midrule
        \texttt{presentation} & Includes scanned images of presentation and overhead slides, transcripts of speeches and statements. Also includes a large amount of press releases. \\
        \midrule
        \texttt{questionnaire} & Includes customer surveys and questionnaires, as well as survey and questionnaire prompts for surveyors. Also includes questionnaires appearing to be part of legal proceedings and investigations. In RVL-CIDP, many questionnaires have a substantial amount of form-like elements.\\
        \midrule
        \texttt{resume} & Includes resumes, curricula vitae (CVs), biographical sketches, executive biographies (e.g., those written in third-person), a small amount of business cards. \\
        \midrule
        \texttt{scientific\_pub.} & Mainly papers and articles from scientific journals and book chapters, but also includes book title pages. Also includes news articles from science newsletters. News articles from science newsletters are very similar to the \texttt{news\_article} category.\\
        \midrule
        \texttt{scientific\_rep.} & Includes bioassay, pathology, and test reports; charts, graphs, and tables; research reports (including progress reports), research proposals, abstracts, paper drafts. Many reports and abstracts bear similarities to scientific publications. Many test result documents are similar to documents in the \texttt{specification} category.\\
        \midrule
        \texttt{specification} & Data sheets (including safety data sheets); product, material, and test specifications. Also includes specification change reports. \\
        \bottomrule
    \end{tabular}}
    \caption{RVL-CDIP categories alongside our descriptions and notes.}
    \label{tab:label-guidelines}
\end{table*}
 
 \section{The RVL-CDIP Dataset}
 
 The RVL-CDIP dataset was introduced in \citet{harley2015icdar} as a benchmark for evaluating image-based classification and retrieval tasks.\footnote{\citet{harley2015icdar} referred to RVL-CDIP as \emph{BigTobacco}.}
 Since then, RVL-CDIP has primarily been used as a document type classification benchmark. RVL-CDIP consists of 400,000 document images distributed across 16 document type categories, listed in Table~\ref{tab:rvl-cdip-categories}.
 Example documents from RVL-CDIP are shown in Figure~\ref{fig:rvlcdip_samples}.
 Documents in RVL-CDIP were sampled from the larger IIT-CDIP Test Collection, which itself is a snapshot of the voluminous Legacy Tobacco Documents Library (LTDL) collection --- at that time, LDTL contained approximately 7 million documents \cite{iit-cdip}.\footnote{The LTDL is now called the Truth Tobacco Industry Documents collection, and is included in the broader Industry Documents Library (IDL) hosted by the UCSF Library: \url{https://www.industrydocuments.ucsf.edu/}. For more background on the LTDL, see \citet{d-lib-document-library} and \citet{digital-archives-2022}.}
 These documents were made publicly available as part of legal proceedings and settlements against several American tobacco and cigarette companies and organizations, and as such, the documents in RVL-CDIP are almost exclusively related to the tobacco industry.\footnote{For more background on the history of the litigation and documents, see \citet{cigarette-papers-1996, decades-of-deceit-1999, digital-archives-2022}.}
 
 Most document images in RVL-CDIP capture the initial page of a document; some common exceptions appear to be charts and tables (these are typically labeled as \texttt{scientific\_report}) as well as presentation slides (labeled as \texttt{presentation}).
 Additionally, almost all of the documents (that contain readable text) are in English, although we did find small amounts of documents in other languages (including German, Dutch, French, Spanish, Portuguese, Italian, Japanese, Chinese, Arabic, and Hebrew) as part of our review.
 Examples of non-English RVL-CDIP samples are displayed in Figure~\ref{fig:non-english_samples} in the Appendix.
 
 There are 320,000 training, 40,000 validation, and 40,000 test samples, but \citet{harley2015icdar} provides no information on how the data was partitioned into these splits, so we assume it was done randomly for each of the 16 document categories.
 \citet{harley2015icdar} report that the 16 categories were chosen, in part, because these categories had ample representation (i.e., at least 25,000 samples) in IIT-CDIP.
 Unfortunately, we are unaware of any published guidelines, criteria, rules, or documentation defining or describing each of the 16 RVL-CDIP categories, nor is it clear who or what provided the initial category labels in IIT-CDIP (nor in LTDL).\footnote{\citet{d-lib-document-library} and \citet{digital-archives-2022} indicate that type labels may have been ascribed to the documents by human workers employed at UCSF's LTDL.}
 Thus, we describe how we developed label guidelines for each
 RVL-CDIP document type category in Section~\ref{sec:label-guidelines} below.
 
 \subsection{Establishing Label Guidelines}\label{sec:label-guidelines}
 The RVL-CDIP dataset does not have a published list of descriptions, rules, or guidelines describing each of the 16 document type categories.
We discuss an extensive analysis from which we develop such guidelines.
 
\begin{figure}
     \centering\scalebox{0.5}{
     \includegraphics{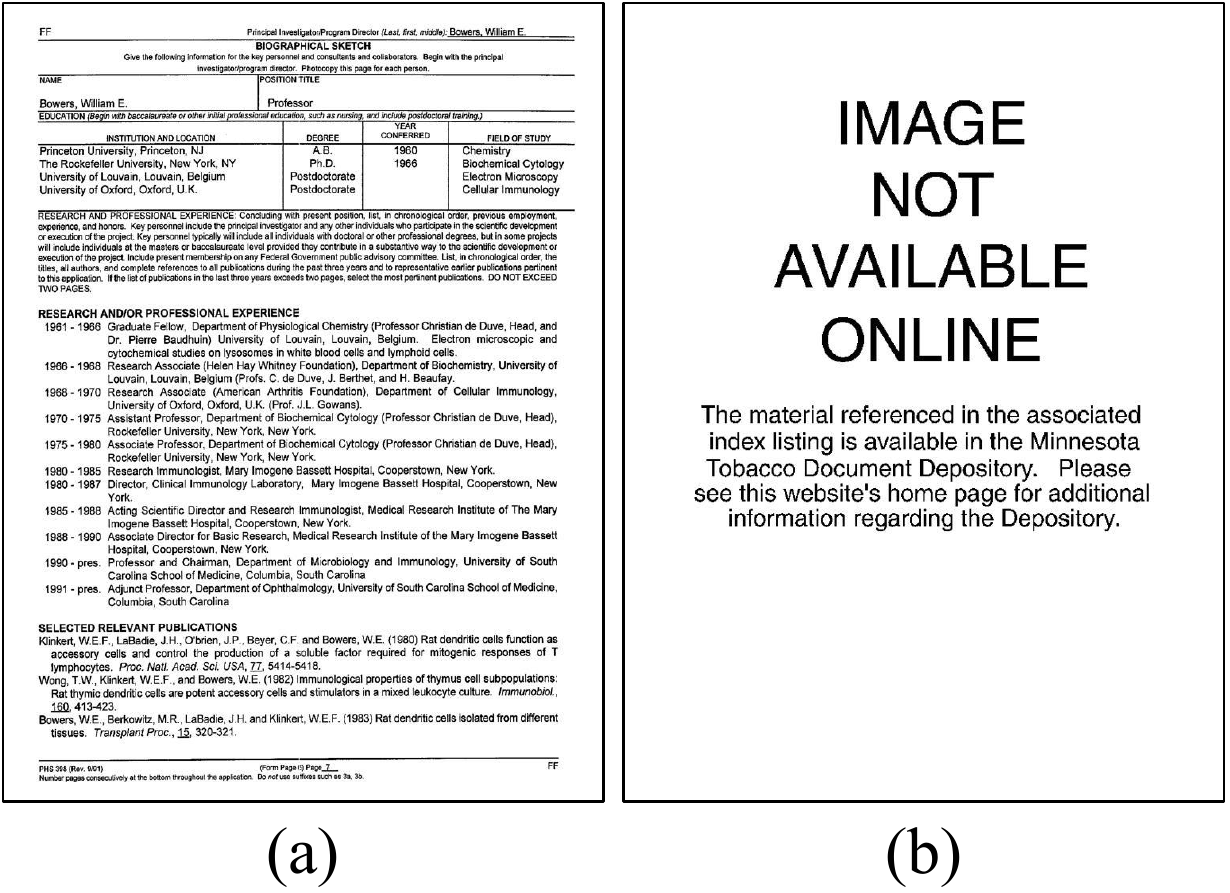}}
     \caption{Example "Biographical Sketch" \texttt{resume} document (a) and "IMAGE NOT AVAILABLE" document found mostly in \texttt{advertisement}.}
     \label{fig:two-examples}
 \end{figure}

 We established our list of guidelines by first sampling 1,000 documents from each of the 16 categories in the training set (for a total of 16,000 documents).
 We then reviewed these samples category by category.
 This review process helped us identify commonalities within each category, and helped us discover that many of the categories seem to have distinct groups of sub-types within them.
 For instance, we found that the \texttt{resume} category is largely composed of (1) resumes and curricula vitae, (2) "Biographical Sketch" documents (i.e., those required for grant applications for the National Institutes of Health (example shown in Figure~\ref{fig:two-examples}a), (3) executive biographies, and (4) scanned business cards.  
 Such cases reveal opportunities for refining and diversifying appropriate categories.
 
 In another category, \texttt{advertisement}, we found samples mostly consisted of advertisements from print-form media like newspapers and magazines, as well as smaller amounts of scripts for television or radio advertisements.
 The \texttt{advertisement} category also included a small amount of document images identical to the one shown in Figure~\ref{fig:two-examples}b.
 We found that this "IMAGE NOT AVAILABLE" document appears mostly in the \texttt{advertisement} category, yet it is an example of a document that we do not include in our label guidelines for this category, as it is not at all faithful to the semantic nature of the \texttt{advertisement} category.

 Our annotation guidelines are listed in Table~\ref{tab:label-guidelines}, along with our notes and observations.
 It was occasionally necessary to review multiple document categories prior to establishing rules.
 \begin{figure*}[t]
     \centering\scalebox{0.49}{
     \includegraphics{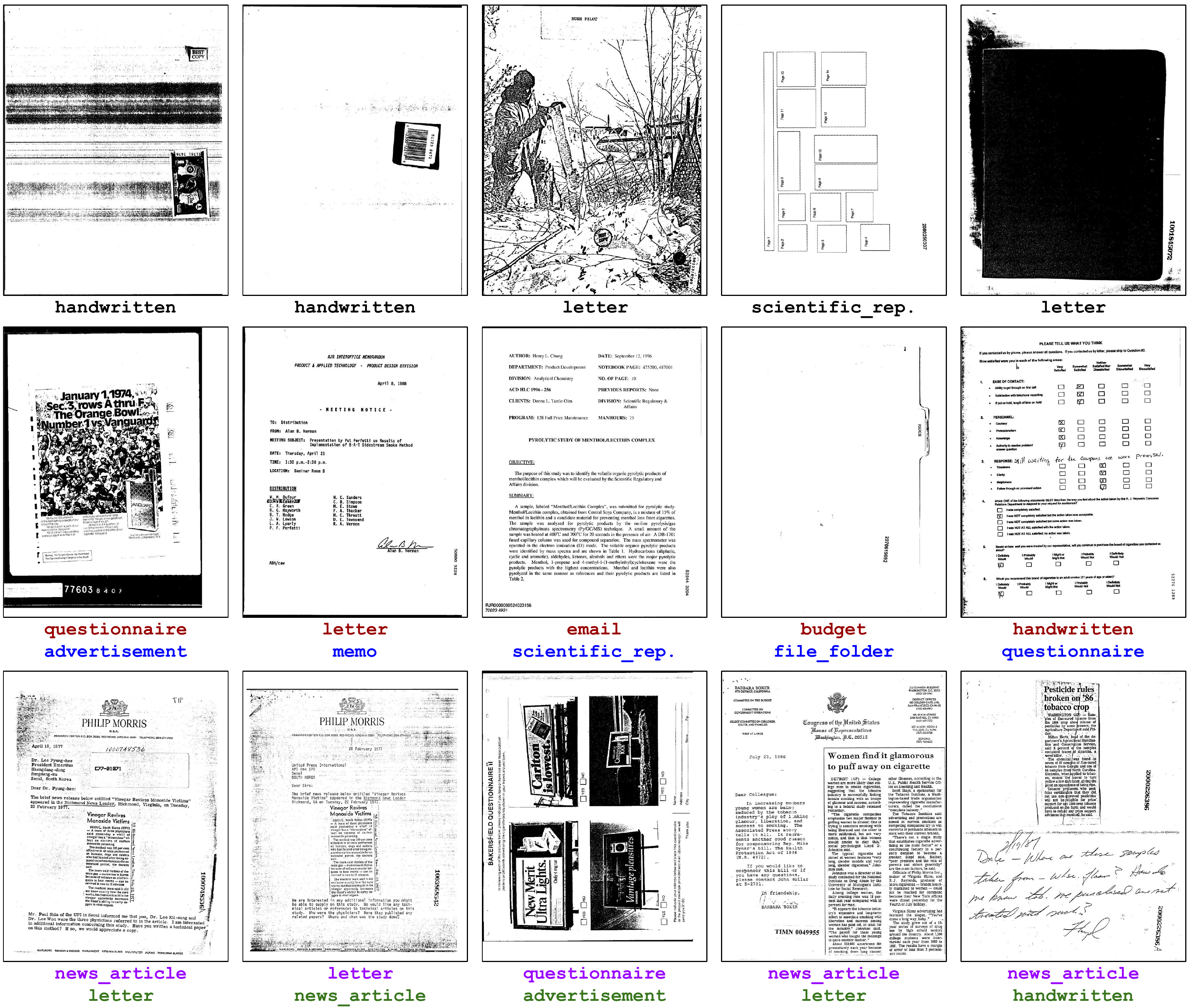}}
     \caption{Example errors and ambiguities. Top row: \emph{unknown}, middle row: \emph{mis-label}, bottom row: \emph{mixed}.}
     \label{fig:error-examples}
 \end{figure*}
 This was the case with the \texttt{budget} and \texttt{invoice} categories, each of which included non-trivial amounts of scanned check images and contribution requests.
 (Examples of cases like these are displayed in Figures~\ref{fig:checks}-\ref{fig:adv} in the Appendix.)
 For cases like these, we annotated these sub-types in the relevant categories in order to estimate their relative frequencies.
 We then would append our annotation guidelines accordingly; for instance, 8.8\% of \texttt{budget} documents and 3.8\% of \texttt{invoice} documents that we reviewed were check images, so our guidelines specify that the \texttt{budget} category consists of check images, while \texttt{invoice} does not.
 Ultimately, our goal with establishing such guidelines is to provide repeatable, verifiable criteria that faithfully reflect the semantic nature of each category. 

 
\section{Label Errors and Ambiguities in RVL-CDIP}\label{sec:label-errors}
 Armed with better knowledge of what constitutes each of the 16 RVL-CDIP categories, we analyze the contents of the RVL-CDIP test set to estimate the amount of label errors and ambiguities found in this set.
 
 We manually checked for errors in the RVL-CDIP test set by sampling 1,000 documents from each of the 16 categories (for a total of 16,000 documents).
 We used our label guidelines established in Section~\ref{sec:label-guidelines} to help us determine the validity of each of these 16,000 samples.
 We tracked several types of errors and ambiguities:
 (1) documents found in a category that clearly are mis-labeled and instead belong in a different RVL-CDIP category --- we refer to this error type as \emph{mis-labeled};
 (2) documents that do not appear to have a single clear RVL-CDIP label --- we refer to this label type as \emph{unknown};
 (3) documents that have mixed or multiple features that belong to at least two RVL-CDIP categories --- we refer to this type as \emph{mixed}.
 Examples of documents exhibiting these error types can be seen in Figure~\ref{fig:error-examples}.
 We point out a particularly interesting \emph{mixed} case: the first two \emph{mixed} examples are nearly identical, but the original label is \texttt{news\_article} in one case but \texttt{letter} in the second.
 More examples are shown in the Appendix in Figures~\ref{fig:more-unknowns}--\ref{fig:more-mixed}.
 
  \begin{table}[]
    \centering\scalebox{0.8}{
    \begin{tabular}{llllll}
    \toprule
    \textbf{Category} & \rotatebox{60}{\emph{\textbf{mis-labeled}}} & \rotatebox{60}{\textbf{\emph{unknown}}} & \rotatebox{60}{\textbf{\emph{Tot. Error}}} & \rotatebox{60}{\textbf{\emph{mixed}}} & \\
    \midrule
        \texttt{advertisement} & 1.9\% & 4.5\% & 6.4\% & 3.5\% & \\
        \texttt{budget} & 9.7\% & 4.1\% & 13.8\% & 1.5\% & \\
        \texttt{email} & 1.7\% & 8.3\% & 10.0\% & 0.4\% & \\
        \texttt{form} & 4.4\% & 6.4\% & 10.8\% & 0.5\% & \\
        \texttt{file\_folder} & 0.4\% & 3.1\% & 3.5\% & 1.9\% & \\
        \texttt{handwritten} & 2.5\% & 5.2\% & 7.7\% & 2.4\% & \\
        \texttt{invoice} & 9.7\% & 1.3\% & 11.0\% & 0.2\% & \\
        \texttt{letter} & 13.5\% & 3.4\% & 16.9\% & 0.5\% & \\
        \texttt{memo} & 2.0\% & 2.4\% & 4.4\% & 2.1\% & \\
        \texttt{news\_article} & 4.6\% & 2.5\% & 7.1\% & 0.4\% & \\
        \texttt{presentation} & 1.8\% & 4.9\% & 6.7\% & 1.0\% & \\
        \texttt{questionnaire} & 5.6\% & 7.3\% & 12.9\% & 6.9\% & \\
        \texttt{resume} & 0.2\% & 1.4\% & 1.6\% & 0.4\% & \\
        \texttt{scientific\_pub.} & 2.5\% & 1.8\% & 4.3\% & 0.0\% & \\
        \texttt{scientific\_rep.} & 4.6\% & 3.9\% & 8.5\% & 5.6\% & \\
        \texttt{specification} & 1.5\% & 1.9\% & 3.4\% & 0.4\% & \\
        \midrule
        Average & 4.2\% & 3.9\% & 8.1\% & 1.7\% \\
    \bottomrule
    \end{tabular}}
    \caption{Estimated label error and multi-label rates in the RVL-CDIP test set.}
    \label{tab:estimated-error-rates}
\end{table}

\begin{figure*}
    \centering\scalebox{0.475}{
    \includegraphics{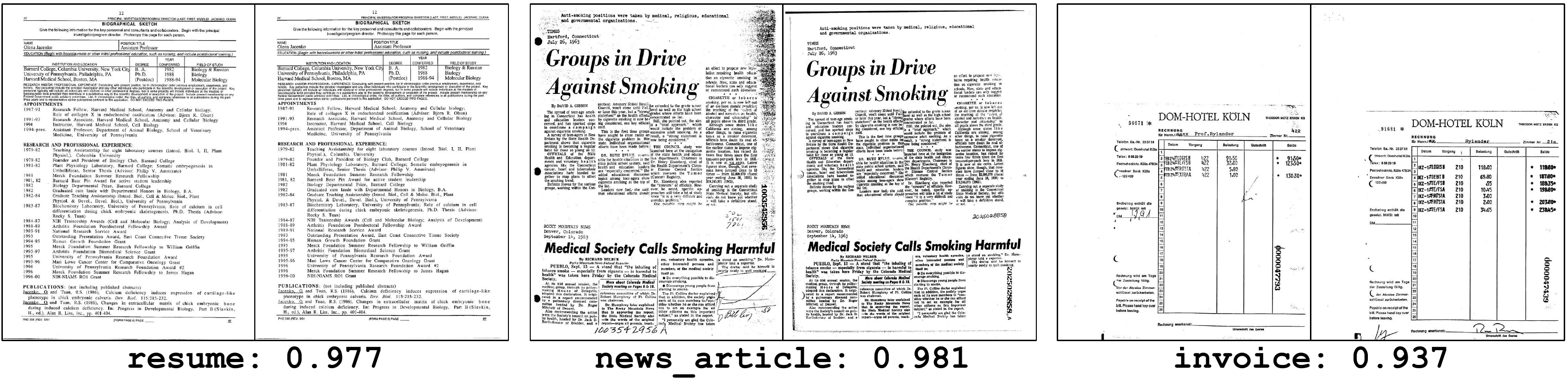}}
    \caption{Example test-train pairs with corresponding maximum cosine similarity scores. These three example pairs show instances of near-duplicates (left and center) and documents that have highly similar structure (right).}
    \label{fig:example-similarity-pairs}
\end{figure*}
 
\paragraph{Findings.}
Our estimated error rates in the RVL-CDIP test set are shown in Table~\ref{tab:estimated-error-rates}.
We estimate that error rates (i.e., combined rates for \emph{mis-labeled} and \emph{unknown}) range between 1.6\% (in the case of \texttt{resume}) and 16.9\% (in the case of \texttt{letter}).
The average of each category's error rates is 8.1\%, which is higher than the classification accuracy error rates reported by many state-of-the-art models listed in Table~\ref{tab:model-scores}.
In some cases, the majority of a category's errors were mis-labels of a particular type.
For instance, about 59\% of the erroneous \texttt{letter} documents we reviewed were actually \texttt{memo} documents.
Similarly, 74\% of the erroneous \texttt{invoice} documents were actually budget documents.
 Lastly, roughly 1.7\% of RVL-CDIP's test set is data that have multiple valid labels.



\section{Overlap Between Test and Train Splits}\label{sec:test-train-overlap}

Our analysis also reveals a substantial degree of undesirable overlap between train and test samples within RVL-CDIP. 
To measure this overlap, we use an approach similar to \citet{larson-etal-2019-outlier} and \citet{elangovan-etal-2021-memorization}, which, for each test sample in each document type category, finds the maximally similar sample in the same document type category's training split.
We then average these maximum similarity scores together for each document category.
That is, for each document category $C$ in RVL-CDIP, we compute 
$$
\frac{1}{|test_C|} \sum_{b\in test_C} \max_{a\in train_C} sim(a,b)
$$
where $a$ and $b$ are samples from category $C$'s train and test splits, respectively.
We use CLIP \cite{clip-model-2021} to extract a 512-dimension feature embedding from each sample, and use cosine similarity for $sim(\cdot,\cdot)$.
We note that this vector-based similarity technique is common practice in the image- and information retrieval (e.g., \citet{neural-codes-2014}).

\begin{table}[]
    \centering\scalebox{0.7}{
    \begin{tabular}{lcc}
    \toprule
    \textbf{Category} & \rotatebox{0}{\textbf{mean}} & \rotatebox{0}{\textbf{median}}\\
    \midrule
        \texttt{advertisement} & 0.893 & 0.903 \\
        \texttt{budget} & 0.963 & 0.968\\
        \texttt{email} & 0.976 & 0.982\\
        \texttt{form} & 0.948 & 0.956\\
        \texttt{file\_folder} & 0.967 & 0.974\\
        \texttt{handwritten} & 0.945 & 0.952\\
        \texttt{invoice} & 0.962 & 0.966\\
        \texttt{letter} & 0.953 & 0.960\\
        \texttt{memo} & 0.957 & 0.961\\
        \texttt{news\_article} & 0.919 & 0.936\\
        \texttt{presentation} & 0.929 & 0.945\\
        \texttt{questionnaire} & 0.961 & 0.968\\
        \texttt{resume} & 0.965 & 0.967\\
        \texttt{scientific\_pub.} & 0.936 & 0.955\\
        \texttt{scientific\_rep.} & 0.950 & 0.961\\
        \texttt{specification} & 0.972 & 0.978\\
        \midrule
        Average & 0.950 & 0.958 \\
    \bottomrule
    \end{tabular}}
    \caption{Mean and median of maximum cosine similarity scores between train and test sets for each RVL-CDIP category.}
    \label{tab:data-similarity-scores}
\end{table}

\paragraph{Findings.}
Average and median of the maximum similarity scores for test-train pairs are shown in Table~\ref{tab:data-similarity-scores} for each RVL-CDIP category.
Overall, we see a high degree of similarity across test and train data: mean scores range between 0.893 (\texttt{advertisement}) and 0.976 (\texttt{email}), with an average of 0.950.
Ten of the 16 document categories have average scores at- or above 0.95.
The median score for each category is larger than the mean in all cases, indicating a long tail in the distribution of scores.
Indeed, we see this in Figure~\ref{fig:hist-similarities} (in Appendix), which charts the distribution of similarity scores for all test data in RVL-CDIP.
Figure~\ref{fig:example-similarity-pairs} shows three examples of test-train pairs with similarity scores ranging between 0.937 and 0.981.
Two of the three pairs in Figure~\ref{fig:example-similarity-pairs} seem to be near-duplicates, where there appear to be minor differences in scanning or noise artifacts between each document.
In the third (\texttt{invoice}) example, we see that the two samples are distinct, yet both share a large degree of similarity because both use the same document template (e.g., invoices from the same company that are structurally and visually similar but that contain different "data").
We show more example pairs in Figures~\ref{fig:appendix-similarities}--\ref{fig:error_similarities} in the Appendix.

\begin{figure}
    \centering\scalebox{0.55}{
    \includegraphics{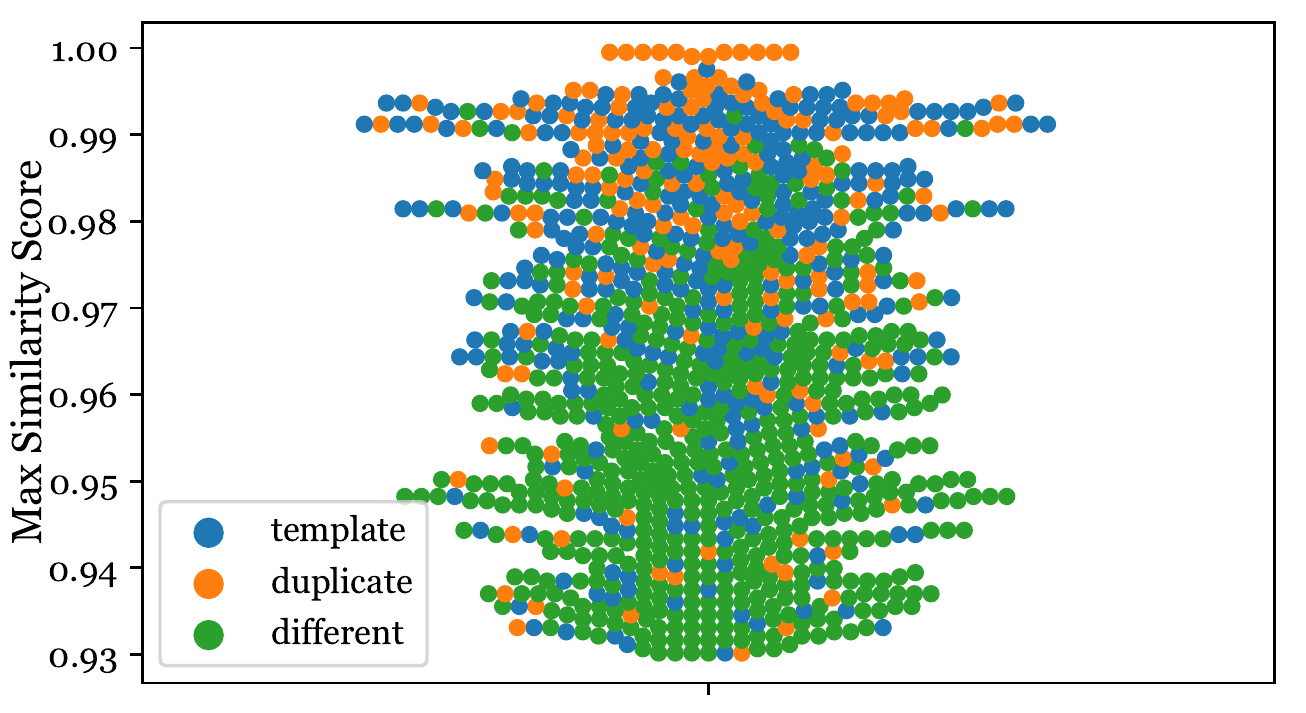}}
    \caption{Sampled subset of maximal similarity scores for test-train pairs with scores between 0.93 and 1.0.}
    \label{fig:similarities-all-swarm}
\end{figure}

To help better understand the similarity scores, we conduct an experiment where we categorize each similarity pair into one of the following: \emph{duplicate}, if the test-train pair represents the same document; \emph{template}, if both documents in a pair use the same document template; and \emph{different}, for all other pairs.
We annotated a sample of 1,086 similarity pairs with maximum similarity scores ranging between 0.93 and 1.0.
A visualization of the relationship between maximal similarity score and match type is shown in Figure~\ref{fig:similarities-all-swarm}, where we observe that the likelihood of a pair being either a \emph{duplicate} or \emph{template} match increases with similarity score.

Considering the overall median maximal similarity score is 0.958, we can estimate a lower-bound for the rate of \emph{duplicate} and \emph{template} match pairs by scaling the proportion of documents above the median maximal score (i.e., half, or 0.5) by the fraction of \emph{duplicate} and \emph{template} matches above the median (0.958). This gives us $0.5 \times 0.641$, and therefore we estimate that at least 32\% of samples from the RVL-CDIP test set have either a duplicate counterpart or a sample that shares a template layout in the training set.
While there is generally no established acceptable number or percentage for test-train overlaps, prior work (e.g., \citet{sogaard-etal-2021-need, elangovan-etal-2021-memorization}) has argued that overlaps are undesirable, and that building generalizable, robust models entails evaluation against novel, unseen data points (e.g., \citet{koh2021wilds,shifts2021,larson-2022-rvlcdip-ood}).

\begin{figure}
    \centering\scalebox{0.41}{
    \includegraphics{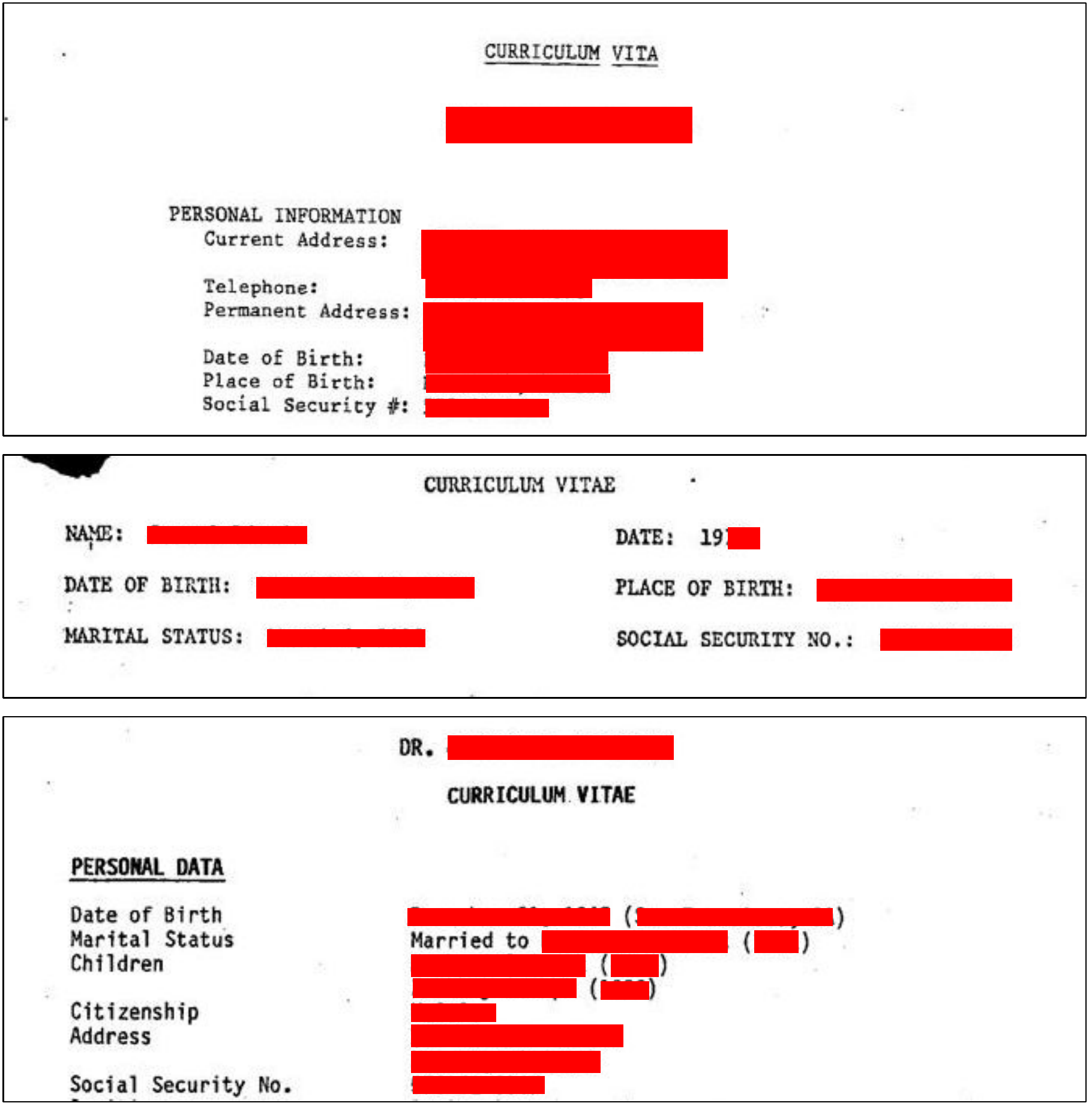}}
    \caption{Example documents from RVL-CDIP showing sensitive personally identifiable information (PII; redacted by us).}
    \label{fig:pii-examples}
\end{figure}

\section{Presence of Sensitive Information}\label{sec:sensitive-information}
While reviewing samples from RVL-CDIP, we noticed that the \texttt{resume} category had a non-trivial quantity of documents that contain sensitive and personally-identifiable entities.
Naturally, resumes typically contain a person's name and basic contact information (e.g., phone numbers or email addresses).
However, we found a plethora of sensitive entities like citizenship and marital statuses, places and dates of birth, names of children and spouses, and national ID numbers like US Social Security and Canadian National ID numbers.

Out of a sample of 1,000 documents from the \texttt{resume} test set, we found that 7.7\% contained a US Social Security Number.
While we recognize that US Social Security numbers were not considered sensitive several decades ago (when many of the \texttt{resume} documents in RVL-CDIP were created), their presence in so many documents in a publicly accessible dataset\footnote{On 9 Feb. 2023, RVL-CDIP tallied "1,765 downloads last month" on the Hugging Face Datasets platform.} is still striking, especially considering the coexistence of this entity type with others like person names, dates and places of birth, etc.
In particular, malicious Social Security numbers are often connected with fraud and identity theft crimes in the USA.
Moreover, the sensitive entities discussed in this section are considered highly sensitive under many state and national laws.\footnote{For example, the State of Michigan's Social Security Number Privacy Act (2004).}
Additionally, we found that 43.6\% of the test resumes contain birth dates, 19.9\% contain places of birth, 11.4\% contain marital (or spousal or parental) statuses, and 8.9\% contain citizenship statuses.
Example documents containing sensitive PII can be seen in Figure~\ref{fig:pii-examples}.

Given the presence of sensitive PII in RVL-CDIP, it is reasonable to wonder if sensitive PII also appears in datasets derived from RVL-CDIP, like FUNSD \cite{funsd}.
Similarly, we also wonder if sensitive PII appears in datasets that were derived from the larger IIT-CDIP or UCSF Industry Documents Library corpora, such as
Tobacco-800 \cite{SignatureDetection-CVPR07, LogoDetection-ICDAR07},
Tobacco-3482 \cite{tobacco-3482},
DocVQA \cite{docvqa-2021},
OCR-IDL \cite{biten2022ocr-idl}, and
TABME \cite{mungmeeprued-etal-2022-tab-this-tabme}.
We will investigate this in future work.

%
%

\section{Discussion and Recommendations}

Given our findings concerning labeling errors, test/train overlap, and presence of sensitive information in the RVL-CDIP document classification benchmark, we discuss several concrete recommendations to raise awareness among researchers engaged in benchmarking classifiers using this dataset:

\textbf{\emph{(0) Sub-Types in RVL-CDIP.}} Our investigation into RVL-CDIP revealed that many of the RVL-CDIP categories are in fact composed of several sub-types.
We encourage researchers and practitioners to be aware of this fact.
For instance, curricula vitae, bigraphical sketches, executive biographies, and business cards are the four sub-types of the \texttt{resume} category.
This finding has implications for modeling tasks where prior knowledge of the label set is assumed, like in zero-shot settings where each category may be specified to the model as a string, as done in \citet{siddiqui-zero-shot-clip}.
Additionally, unsupervised clustering analyses like \citet{finegan-dollak-verma-2020-layout} may exhibit low performance scores on RVL-CDIP due to many of the categories having distinct and disparate sub-types (e.g., radio scripts versus print advertisements in the \texttt{advertisement} category, or business cards versus biographical sketches in the \texttt{resume} category).

\textbf{\emph{(1) Errors.}} Users of RVL-CDIP should be aware that there are many label errors and noisy samples with unknown labels in RVL-CDIP.  Recall from Section~\ref{sec:label-errors} that an estimated 8.1\% of test samples from RVL-CDIP contain label errors, with an additional 1.7\% being ambiguous mixed or multi-label cases. This is problematic for benchmarking new models, since the estimated label error rate is now greater than state-of-the-art model accuracy error rates. Here, the implication is that high-capacity models like CNNs and transformers are now overfitting to noise.
This is indeed the case for models like DiT \cite{li2022dit}, which predict the "IMAGE NOT AVAILABLE" document to be an \texttt{advertisement} document due to its relative abundance in that category's training set.

\textbf{\emph{ (2) Ambiguities.}} Users of RVL-CDIP should be aware that there are many samples in RVL-CDIP that could have multiple valid document type labels. We estimate this number to be 1.7\% of the RVL-CDIP test set. Like label errors, such mixed or multi-label cases make it challenging to evaluate a model effectively, as there are samples for which a model may make a wrong prediction according to the RVL-CDIP test label annotations, but in reality many of these wrong predictions could actually be reasonable.

\textbf{\emph{ (3) Test-Train Overlap.}} Practitioners and researchers should be aware that there is a high degree of overlap between the RVL-CDIP test set and the train set.  Recall from Section~\ref{sec:test-train-overlap} that almost a third of RVL-CDIP test samples have a near-duplicate in the training set for the same document type category, or a training sample that uses the same document template.
This is undesirable, as testing models on data that is very similar to the training data can lead to "inflated" accuracy scores \cite{elangovan-etal-2021-memorization, sogaard-etal-2021-need}.
Moreover, highly similar train and test splits do not facilitate the evaluation of a model's ability to generalize well to new in-domain data.

\textbf{\emph{ (4) Sensitive Information.}} There is an unsettling amount of sensitive information in the RVL-CDIP dataset, which naturally leads to information and data privacy concerns.
We estimate that 7.7\% of \texttt{resume} test samples contain Social Security numbers. While RVL-CDIP is already publicly available, researchers and practitioners should take care when disseminating samples or copies of RVL-CDIP.
Moreover, we highlight that prior work (e.g., \citet{carlini-2021-extracting}) showed that it is possible to extract training data from machine learning models, making production deployments of models trained on RVL-CDIP an information privacy and security risk.

\textbf{\emph{Suggestions for a future dataset.}}
We suggest the development and adoption of a new benchmark for evaluating document classifiers.
Several qualities of a such a benchmark would include (1) minimal label errors; (2) multi-label annotations, to allow for modeling more natural occurrences of documents; (3) minimal test-train overlap; (4) absence of sensitive information.
Going beyond the points made in this paper, a new benchmark would do well to be (5) large-scale, consisting of 100+ or even 250+ document categories, to test a model's ability to handle breadth, and (6) multi-lingual, to benchmark language transfer approaches.

\section{Conclusion}
RVL-CDIP has been used as the \emph{de facto} benchmark for evaluating state-of-the-art document classification models, but
this paper provides an in-depth analysis of the RVL-CDIP dataset and shows that there are several undesirable characteristics of this dataset.
We first provide a set of label guidelines for each RVL-CDIP category, and we use this to help us quantify the presence of errors in RVL-CDIP, finding that the RVL-CDIP test set contains roughly 8.1\% label errors.
We then observe that roughly a third of the test data is highly similar to the training set. Lastly we observe an unsettling amount of personally sensitive information in RVL-CDIP.
Given these findings, we offer suggestions for a new document classification benchmark.

\section*{Limitations}
The RVL-CDIP dataset has no official set of label guidelines, making error analyses challenging since we could not rely on pre-defined rules.
For this reason we followed best practices to create annotation rules to help us in our error analysis.
Detecting duplicates in RVL-CDIP is also challenging, as two documents may appear to be the same, but may have minor differences due to scanning artifacts or even different indexing labels (it appears that many of the documents have been scanned and included in IIT-CDIP more than once).
Therefore we again have to rely on best judgement when labeling pairs as duplicates (or near-duplicates).
Additionally, due to limitations in human resources, we were unable to exhaustively inspect all 400,000 RVL-CDIP samples for the presence of errors, ambiguities, sensitive information, etc., and thus had to rely on sampling the dataset in order to draw conclusions.

\section*{Acknowledgements}
We thank Nicole Cornehl Lima, Ramla Alakraa, Zongyi Liu, Junjie Shen, Temi Okotore for help with data review, as well as the University of Michigan's Undergraduate Research Opportunity Program (UROP) for their support of these student researchers as well as support for Gordon.
We also thank the anonymous EACL reviewers for their feedback.

\bibliography{acl2020}
\bibliographystyle{acl_natbib}

\newpage\clearpage
\appendix

\section{Appendix}
This appendix is used to provide supplementary material.
Appendix~\ref{sec:cleanlab} discusses using an automated label error dection tool called Cleanlab, and why we ultimately did not use it to aid us in our review of RVL-CDIP.
Appendix~\ref{sec:appendix} provides supplementary visualizations in support of the main paper.
Finally, Appendix~\ref{sec:data-github} details where to find the label annotations developed in this paper.

\begin{figure}
    \centering\scalebox{0.85}{
    \includegraphics{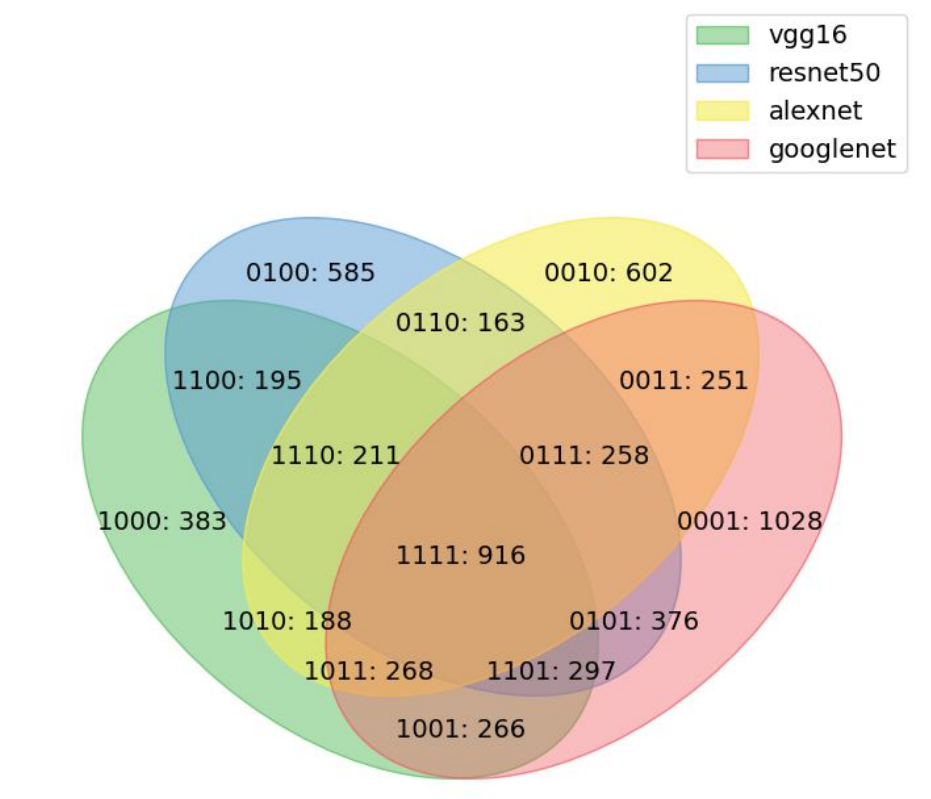}}
    \caption{Venn diagram showing overlaps among predicted label errors in RVL-CDIP's test set using Cleanlab with four different classifier models. The labels are one-hot encoded in this figure (e.g., "\texttt{1010}" indicates the intersection between the Cleanlab predictions for VGG-16 and AlexNet).}
    \label{fig:cleanlab-venn}
\end{figure}

\begin{figure}
    \centering\scalebox{0.5}{
    \includegraphics{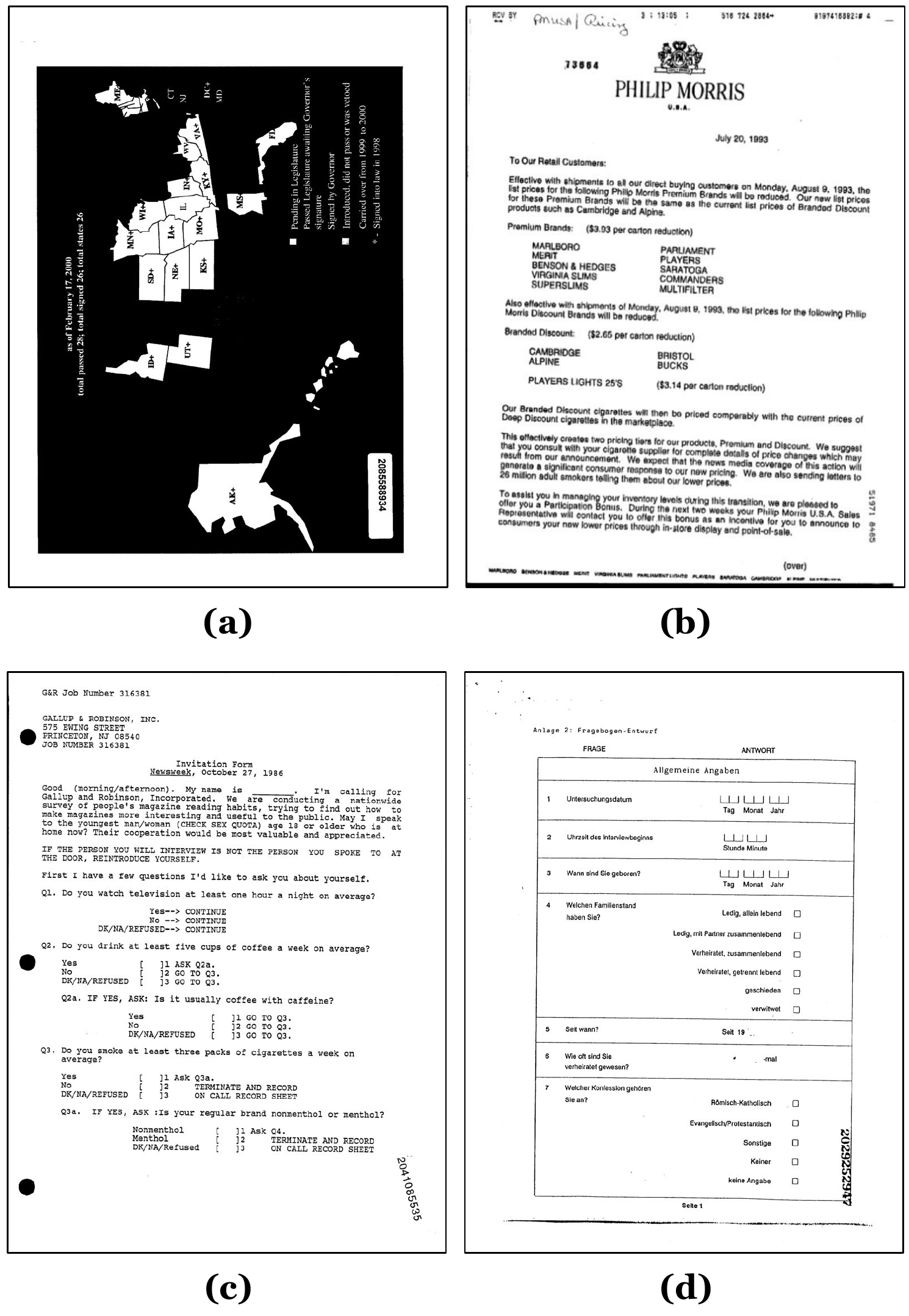}}
    \caption{Problematic label error predictions from Cleanlab with VGG-16.}
    \label{fig:cleanlab-errors}
\end{figure}

\subsection{Cleanlab Discussion}\label{sec:cleanlab}
Automated tools exist for detecting label errors in classification datsets.
One such exemplary tool is Cleanlab, which uses confident learning algorithms to predict label errors in datasets \cite{northcutt2021confidentlearning, test-errors-2021}.
We used the off-the-shelf version of Cleanlab,\footnote{\url{https://github.com/cleanlab/cleanlab}} which aims to identify label errors in a dataset given a model's predictions on that dataset. That is, Cleanlab uses the original data labels, the model's predicted labels, and the model's confidence scores to make a prediction of \emph{error} or \emph{not-error} for each sample.

We used Cleanlab on the RVL-CDIP test set (using models trained on the full RVL-CDIP training set from \citet{larson-2022-rvlcdip-ood}) with four different classifier models: GoogLeNet, AlexNet, ResNet, and VGG-16 (each uses the architecture from \citet{googlenet}, \citet{alexnet}, \citet{res-net-50}, and \citet{vgg-16}, respectively). 
One initial observation was that there was not a "tight" amount of agreement across all four runs of the Cleanlab tool.
This is visualized in Figure~\ref{fig:cleanlab-venn}, where we see that only 916 out of 5,987 RVL-CDIP test samples were predicted as errors by all four runs of Cleanlab.

We also observed that many of the label error predictions made by Cleanlab were themselves problematic.
For instance, Figures~\ref{fig:cleanlab-errors}a and \ref{fig:cleanlab-errors}b show cases where Cleanlab incorrectly predicted a label error: Figure~\ref{fig:cleanlab-errors}a is a valid \texttt{presentation} document, and  Figure~\ref{fig:cleanlab-errors}b is a valid \texttt{letter} document. Figures~\ref{fig:cleanlab-errors}c and \ref{fig:cleanlab-errors}d show \emph{ambiguous} cases where Cleanlab predicted a label error: Figure~\ref{fig:cleanlab-errors}c shows a \texttt{form} document with questionnaire-like elements, while Figure~\ref{fig:cleanlab-errors}d shows a \texttt{questionnaire} document with form-like elements.
However, we argue that these false-positives are not entirely due to the Cleanlab tool, but instead due to the noisy nature of the RVL-CDIP training set: since Cleanlab uses model predictions, and since those models were trained on noisy data, the Cleanlab predictions are therefore bound to be imperfect.
Similarly, we posit that the large amount of test-train overlap leads to brittle models, which also leads to imperfect predictions by Cleanlab.
Indeed, Cleanlab's documentation warns that "Cleanlab performs better if the [model confidence scores] from your model are out-of-sample"\footnote{\url{https://docs.cleanlab.ai/stable/index.html}} and we have argued in the main paper above that high amounts of test-train overlap lead to fewer test cases that are out-of-sample.   

\subsection{Supplementary Visualizations}
\label{sec:appendix}
\begin{figure}
    \centering\scalebox{0.475}{
    \includegraphics{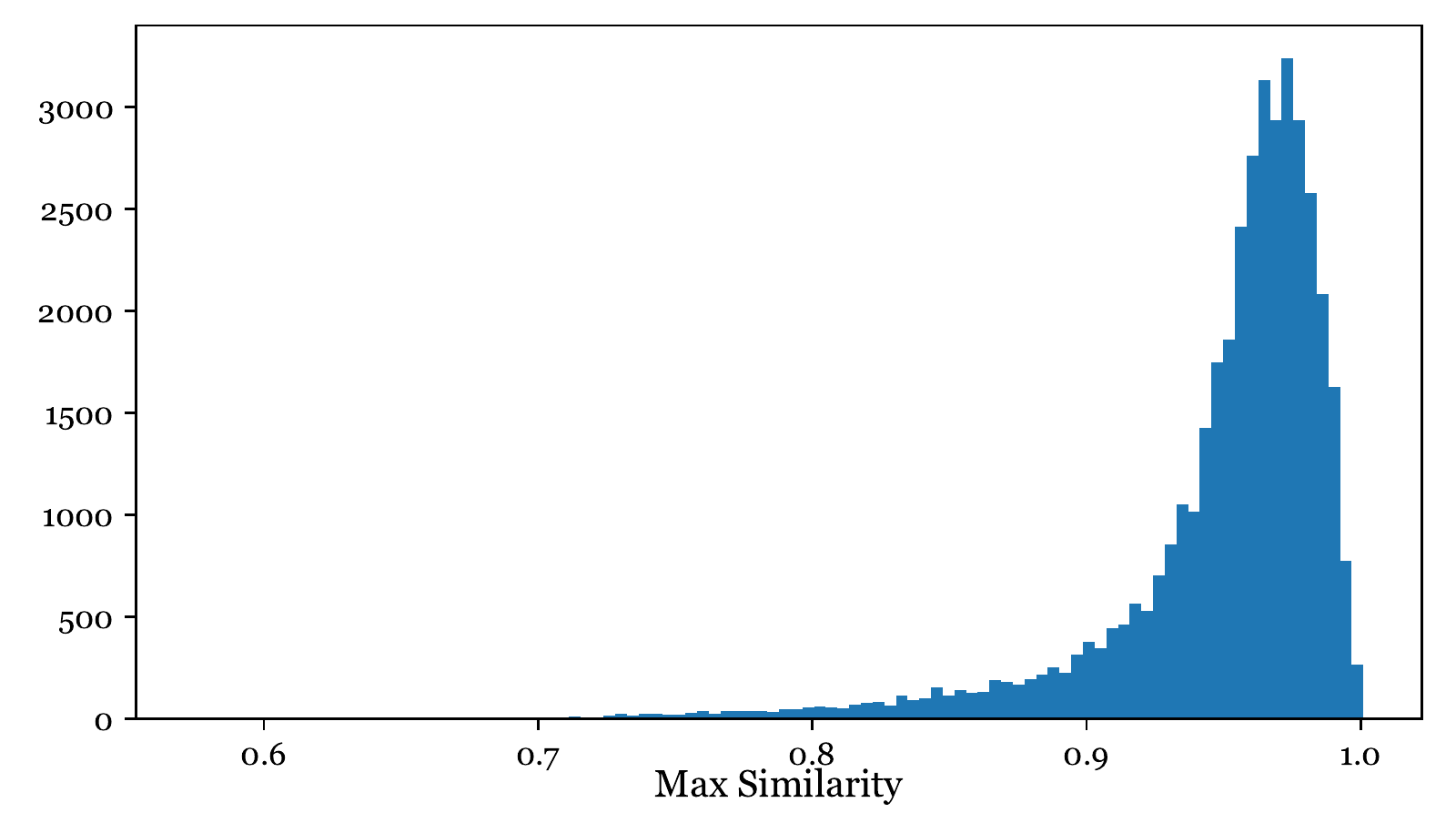}}
    \caption{Maximum similarities between test and train samples for RVL-CDIP test data.}
    \label{fig:hist-similarities}
\end{figure}

\begin{figure*}
    \centering\scalebox{0.49}{
    \includegraphics{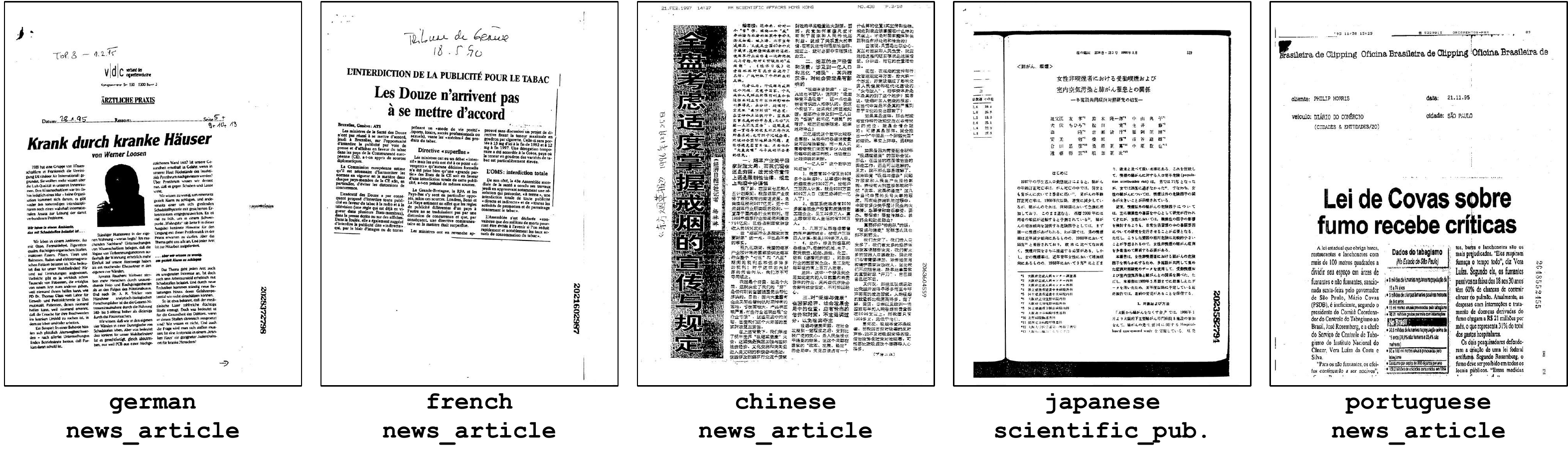}}
    \caption{Example non-English samples from RVL-CDIP.}
    \label{fig:non-english_samples}
\end{figure*}

\begin{figure*}
    \centering\scalebox{0.49}{
    \includegraphics{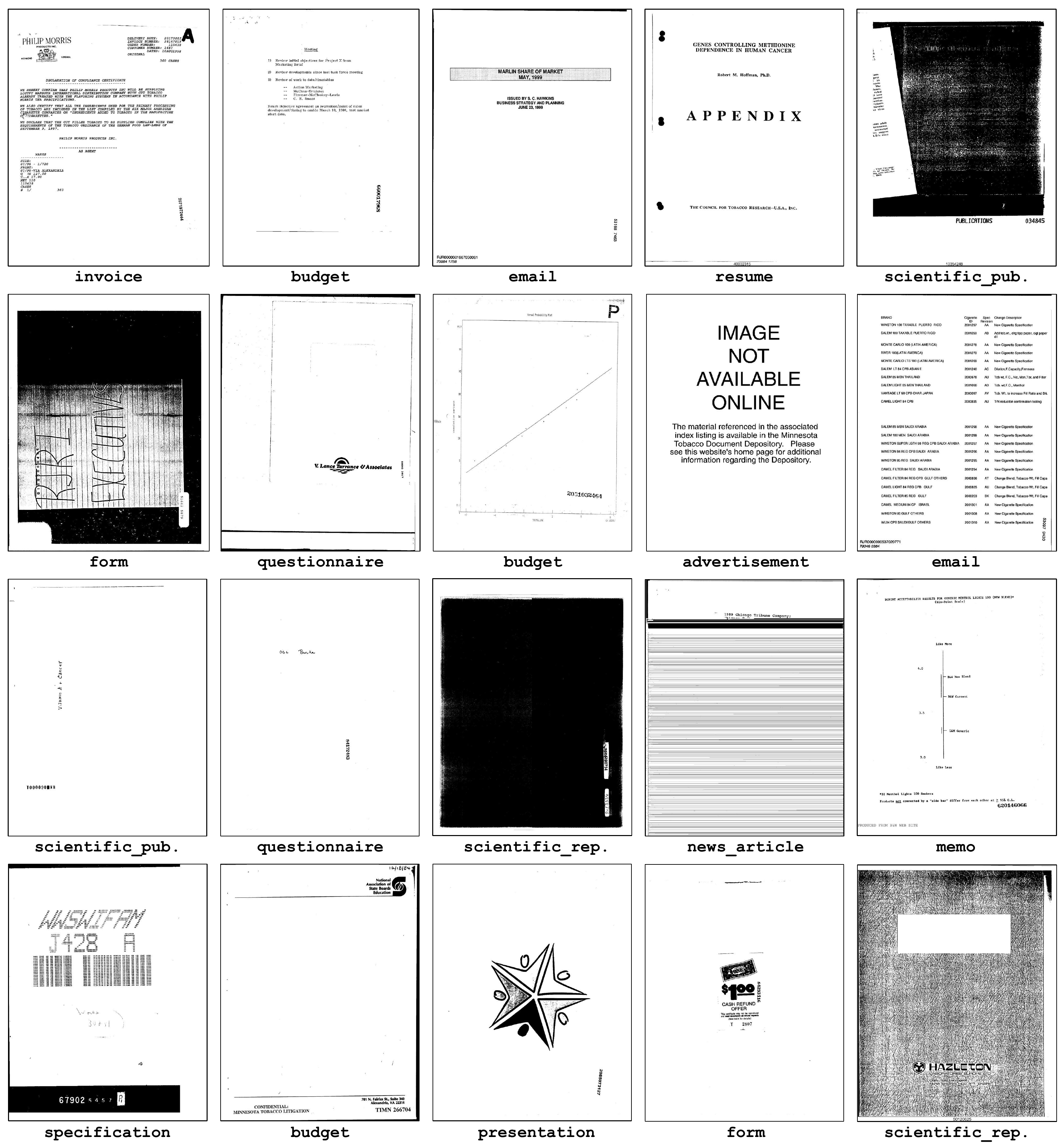}}
    \caption{Examples of \emph{unknown} label errors with corresponding original RVL-CDIP labels.}
    \label{fig:more-unknowns}
\end{figure*}

\begin{figure*}
    \centering\scalebox{0.49}{
    \includegraphics{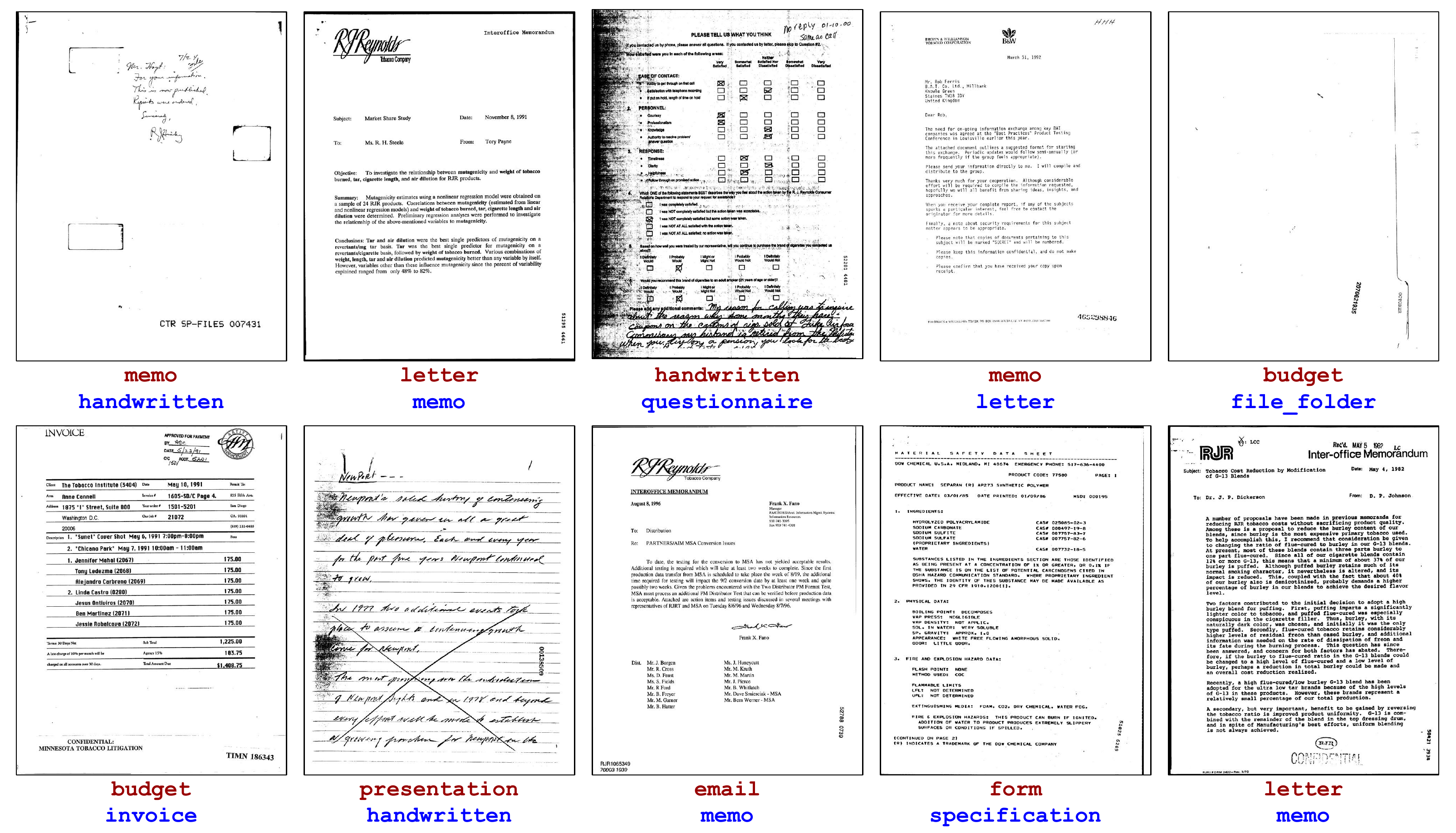}}
    \caption{Examples of \emph{mis-label} label errors with corresponding original (top) and corrected (bottom) RVL-CDIP labels.}
    \label{fig:more-mislabels}
\end{figure*}

\begin{figure*}
    \centering\scalebox{0.49}{
    \includegraphics{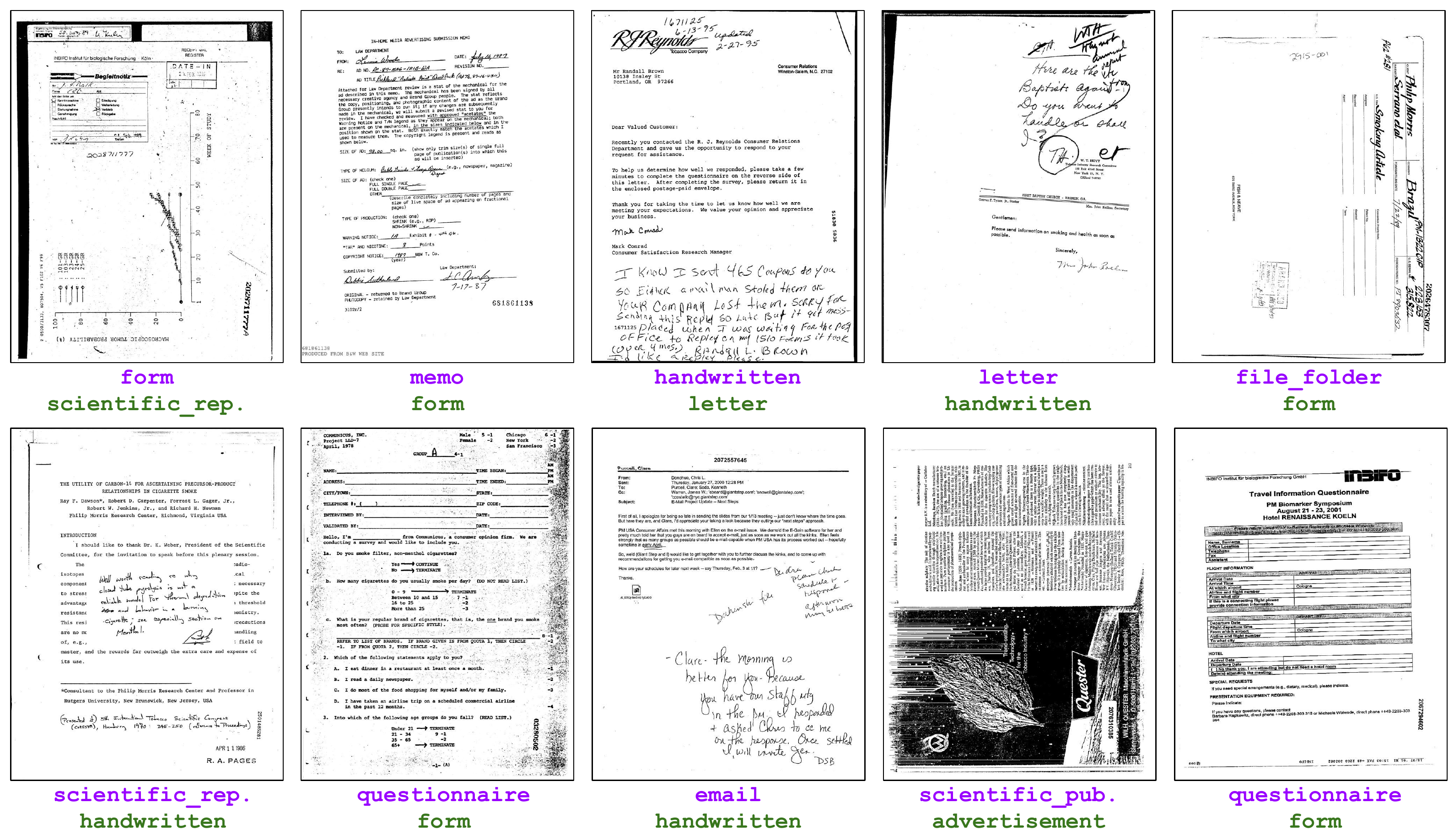}}
    \caption{Examples of mixed or multi-label documents from RVL-CDIP. The original RVL-CDIP label is shown first (top) and the additional valid label second (bottom).}
    \label{fig:more-mixed}
\end{figure*}

\begin{figure*}
    \centering\scalebox{0.47}{
    \includegraphics{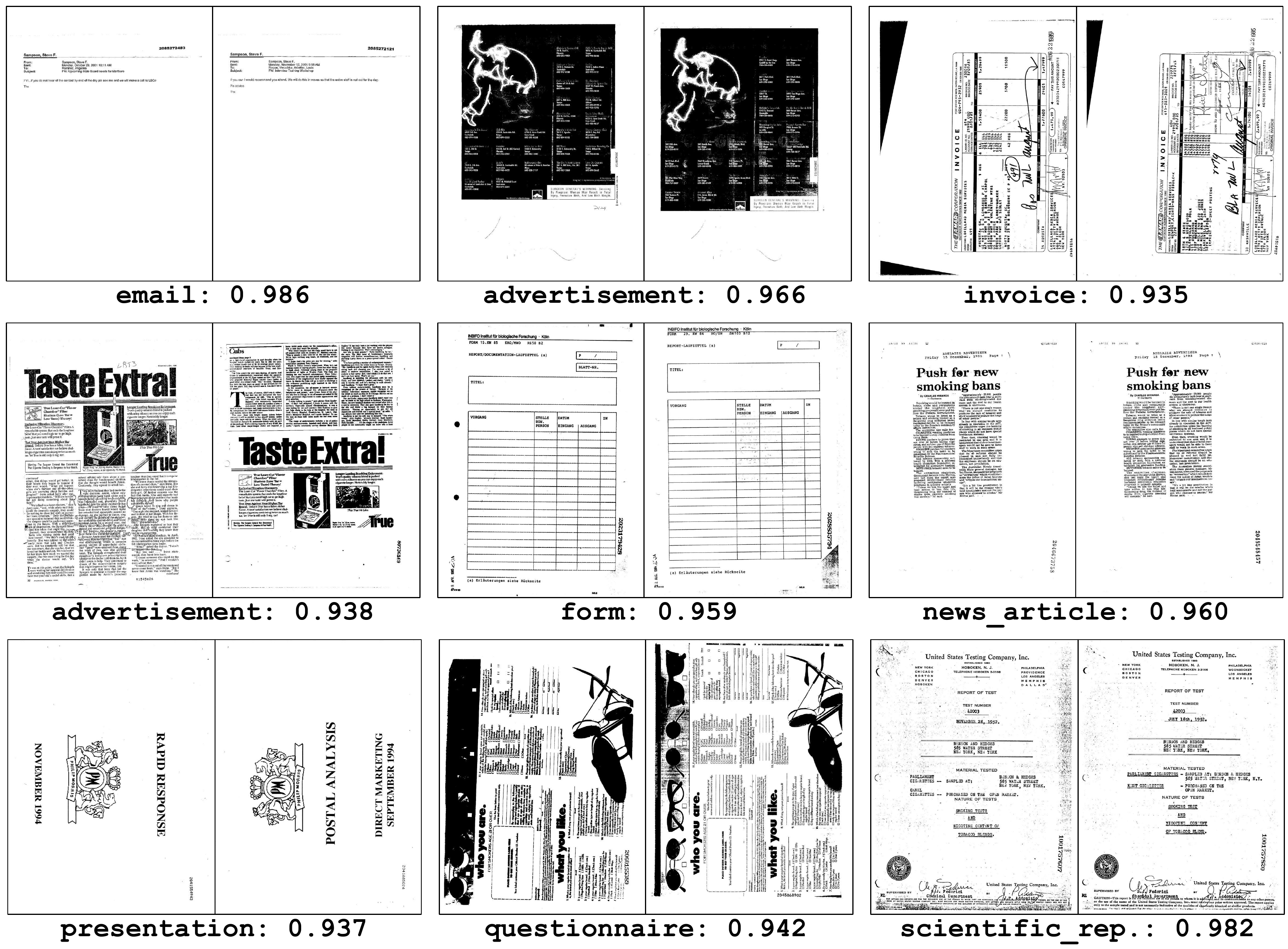}}
    \caption{Examples of test-train pairs with corresponding cosine similarity scores.}
    \label{fig:appendix-similarities}
\end{figure*}

\begin{figure*}
    \centering\scalebox{0.47}{
    \includegraphics{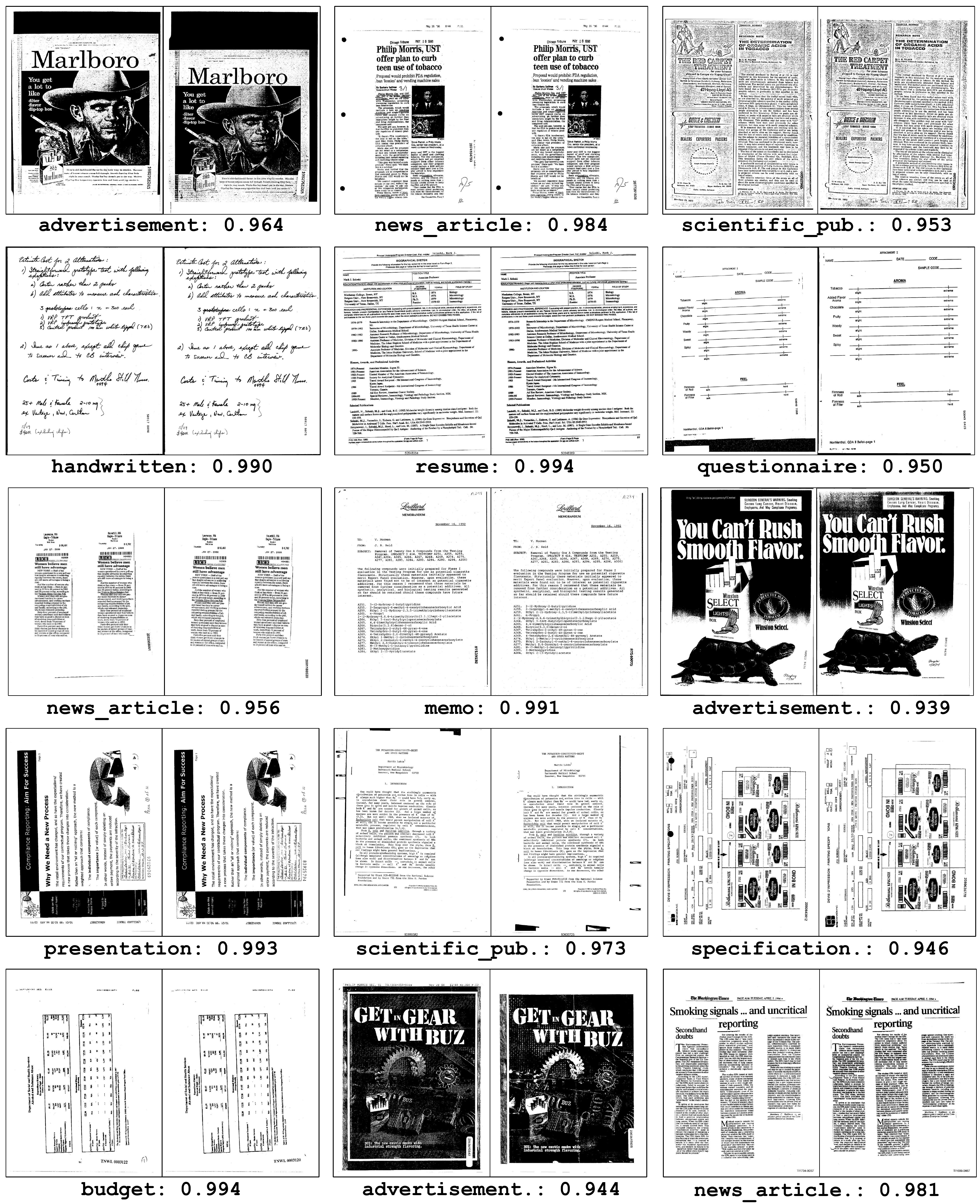}}
    \caption{Example test-train duplicate pairs.}
    \label{fig:example_duplicates}
\end{figure*}

\begin{figure*}
    \centering\scalebox{0.47}{
    \includegraphics{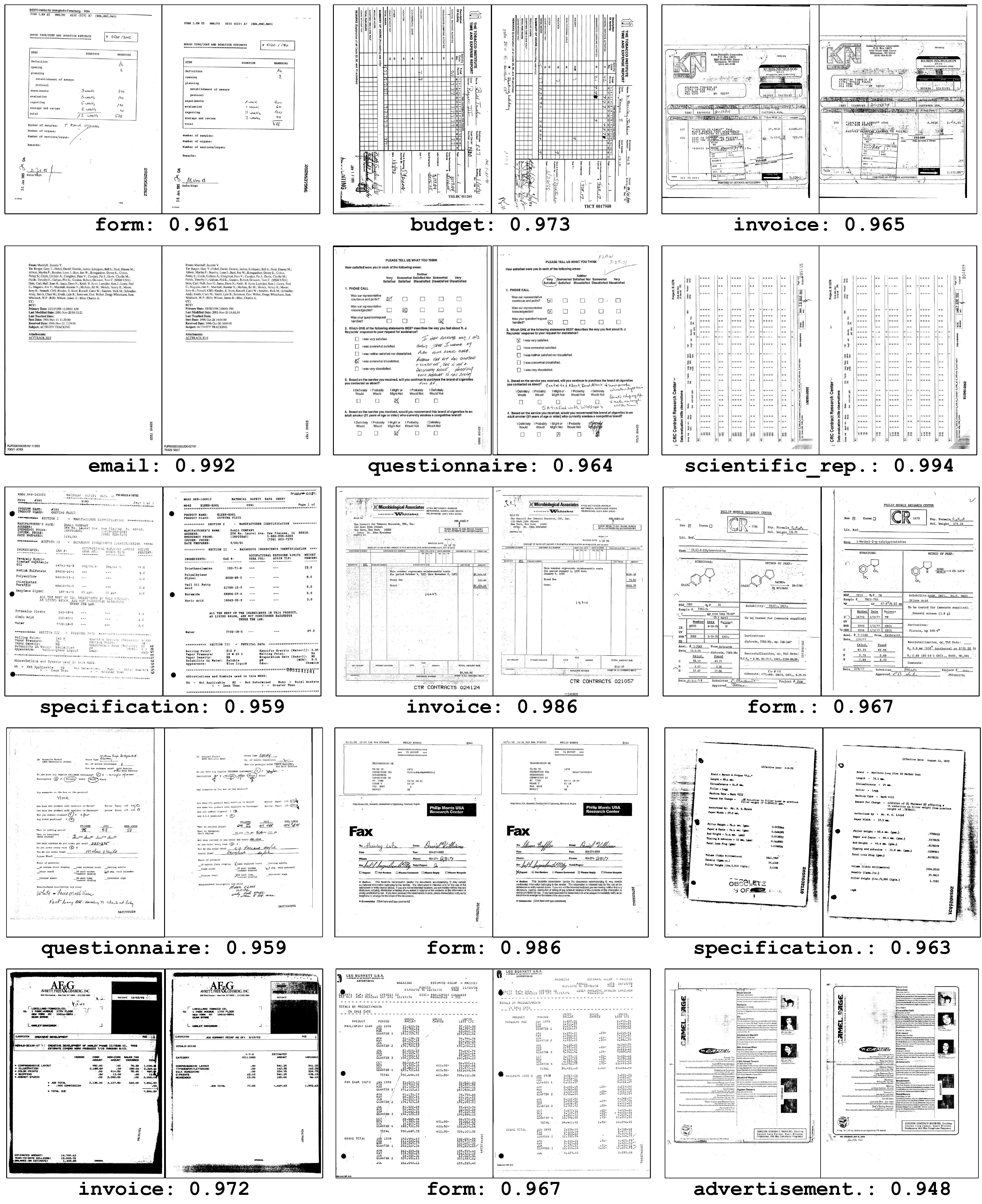}}
    \caption{Example test-train pairs with a high level of similarity due to overlap in document templates.}
    \label{fig:example_template_similarity}
\end{figure*}

\begin{figure*}
    \centering\scalebox{0.47}{
    \includegraphics{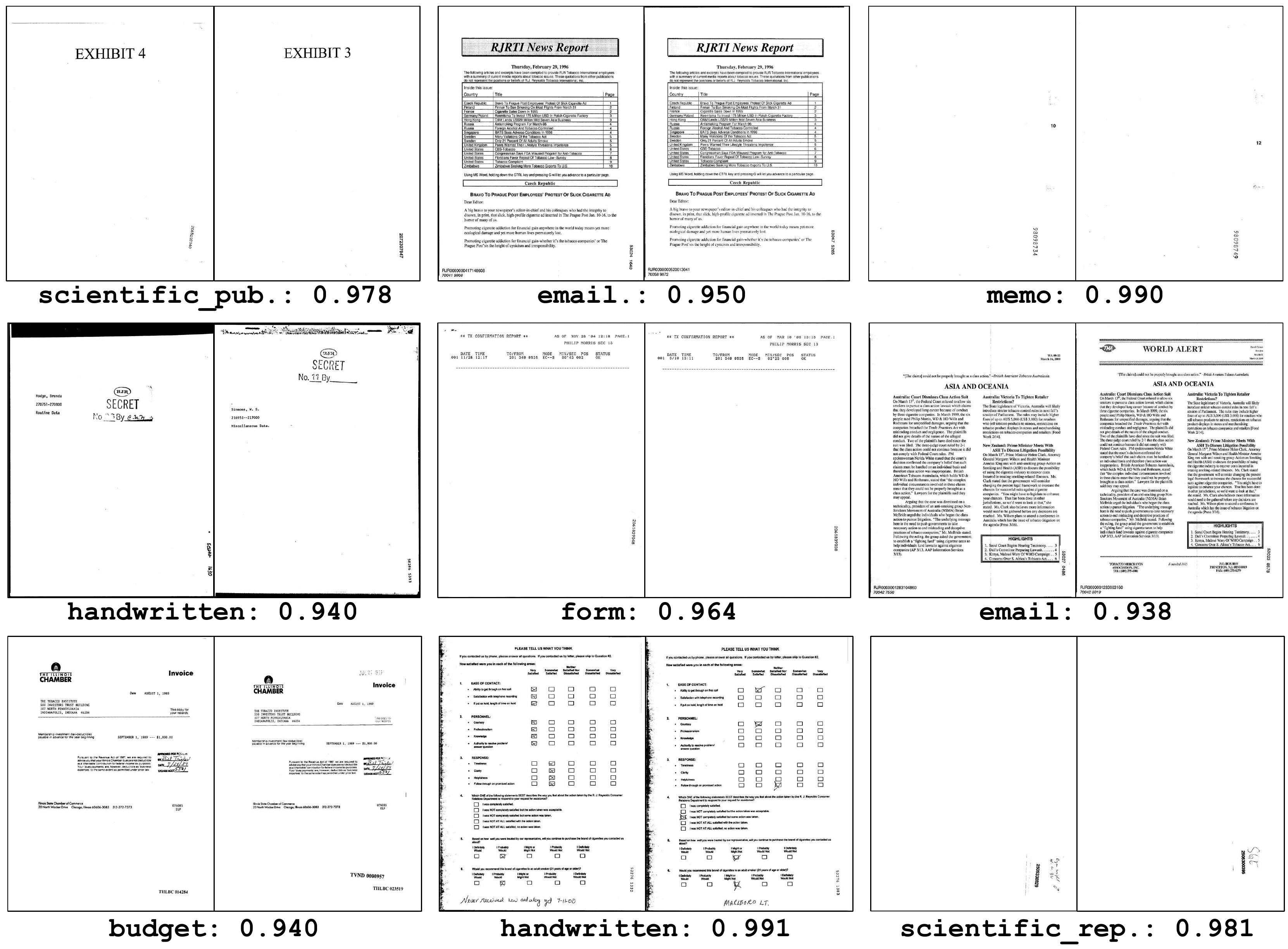}}
    \caption{Example test-train pairs that have erroneous labels.}
    \label{fig:error_similarities}
\end{figure*}

\begin{figure*}
    \centering\scalebox{0.5}{
    \includegraphics{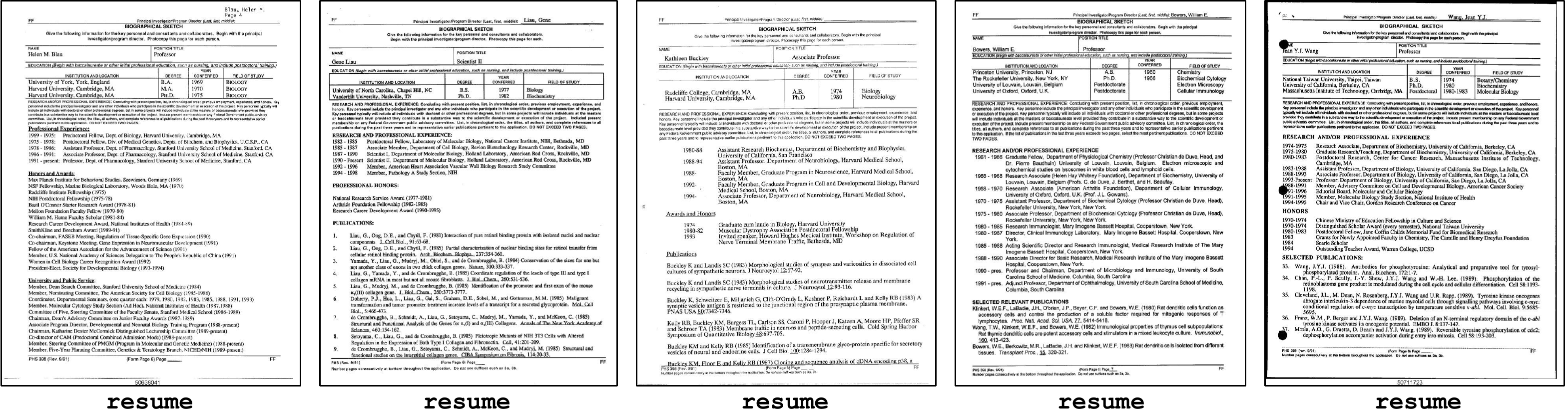}}
    \caption{Examples of "Biographical Sketch" documents, which are abundant in the \texttt{resume} category.}
    \label{fig:biosketches}
\end{figure*}

\begin{figure*}
    \centering\scalebox{0.5}{
    \includegraphics{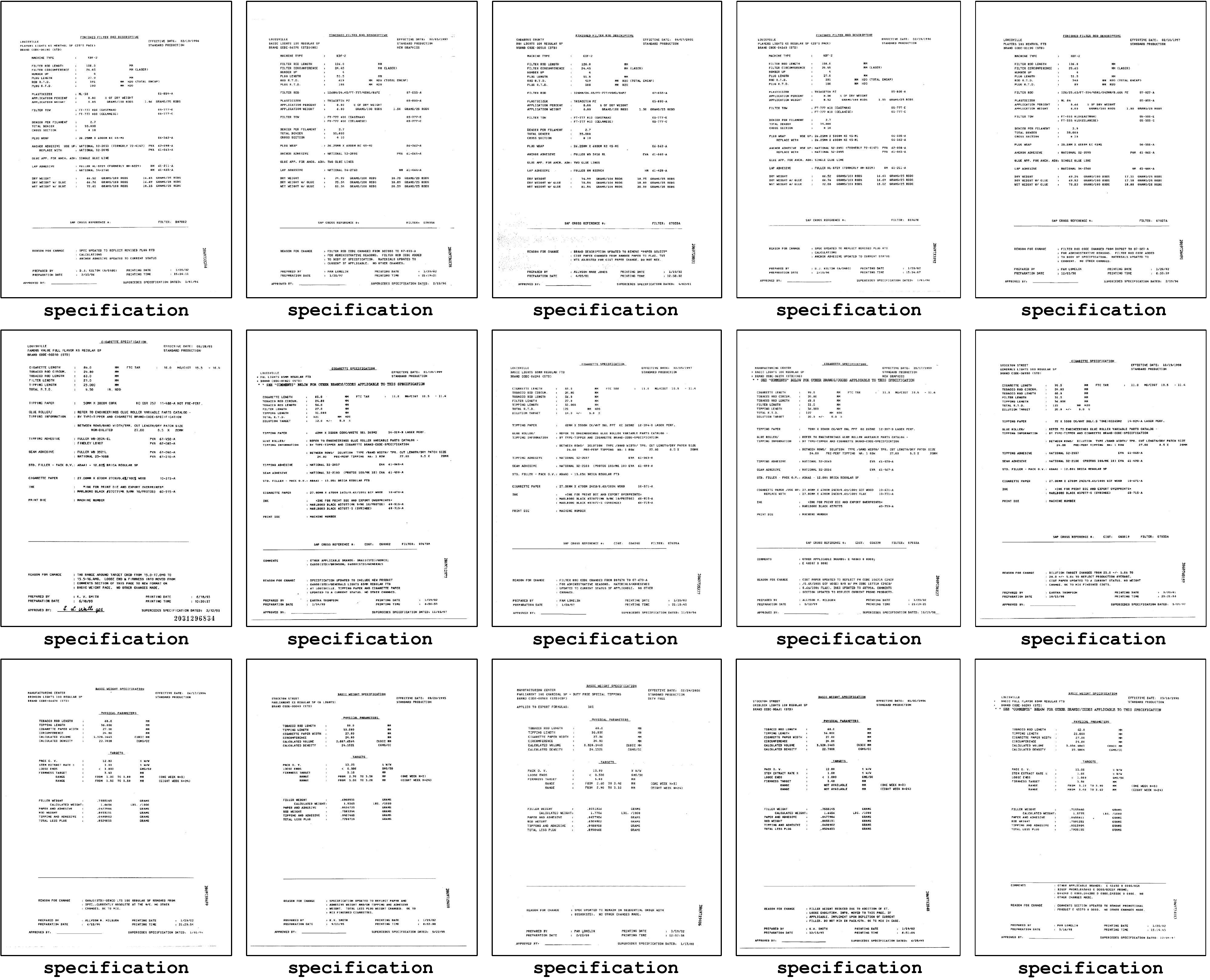}}
    \caption{Examples of three common types of documents within the \texttt{specification} document categor: "Finished Filter Rod Descriptive" documents (top row), "Cigarette Specification" (middle row), "Basic Weight Specification" (bottom row).}
    \label{fig:specification_templates}
\end{figure*}

\begin{figure*}
    \centering\scalebox{0.5}{
    \includegraphics{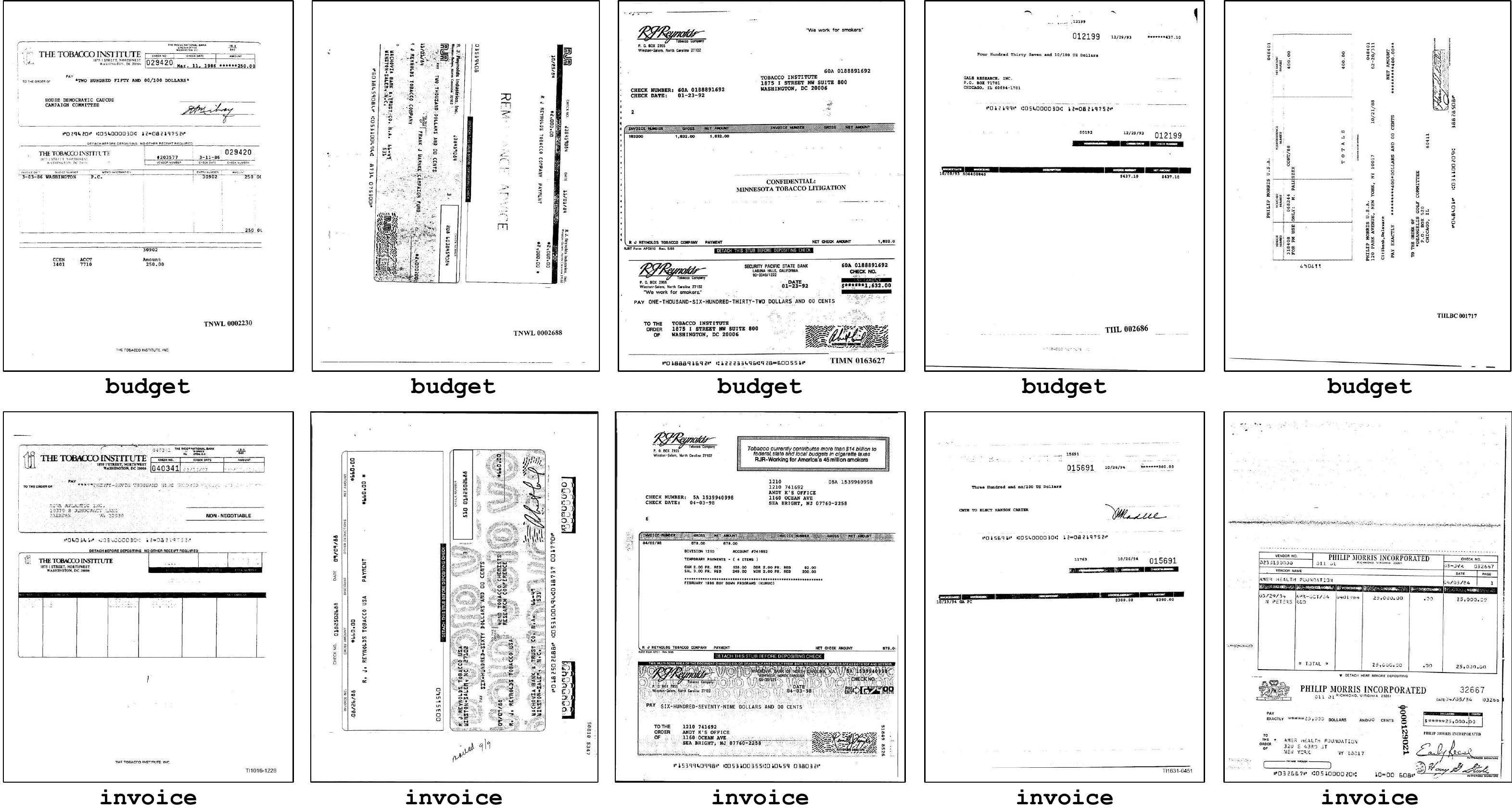}}
    \caption{Examples of check images from the \texttt{budget} (top) and \texttt{invoice} document categories.}
    \label{fig:checks}
\end{figure*}

\begin{figure*}
    \centering\scalebox{0.5}{
    \includegraphics{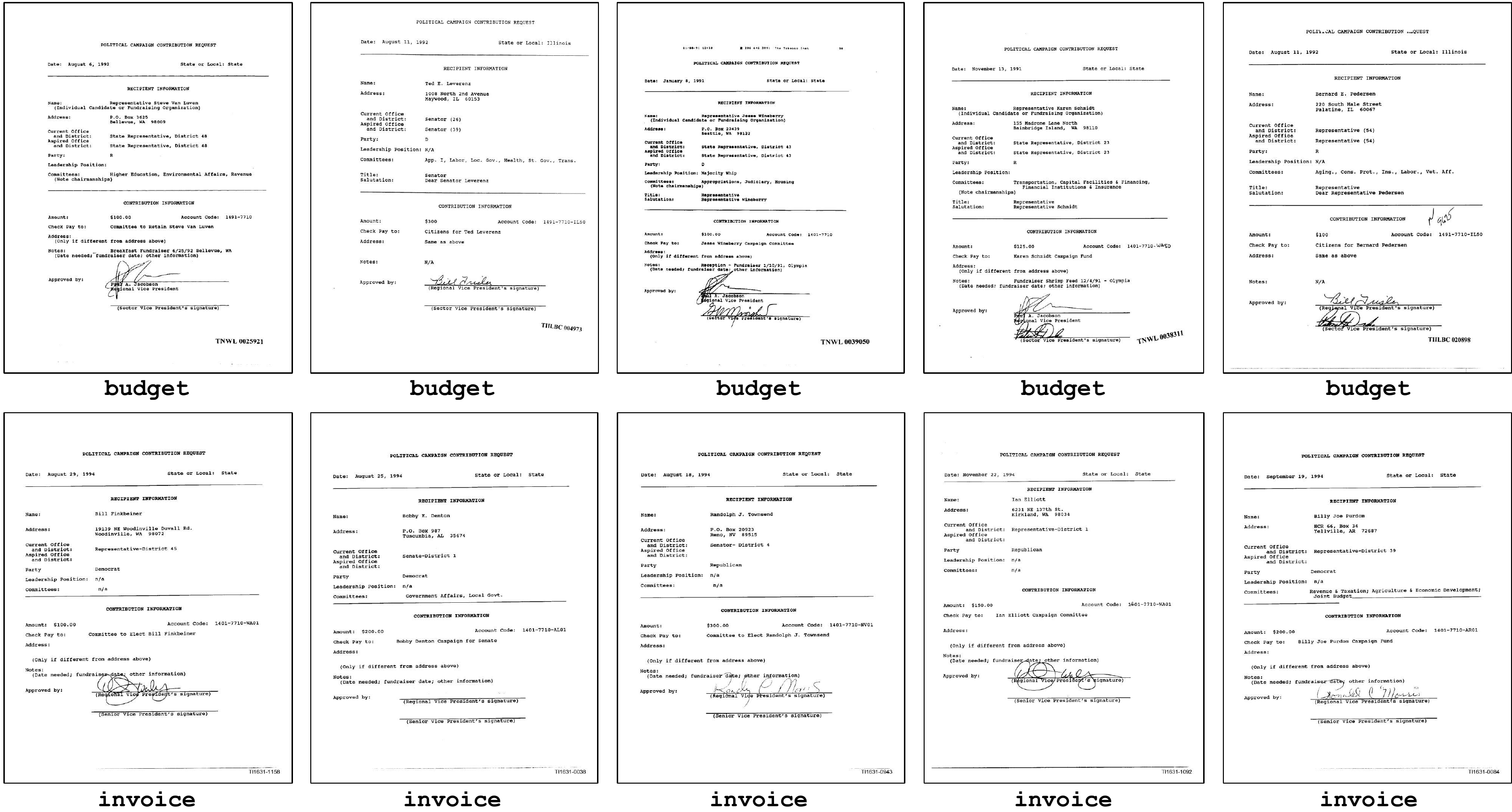}}
    \caption{Examples of "Political Campaign Contribuiton Request" documents from \texttt{budget} (top row) and \texttt{invoice} (bottom row) categories.}
    \label{fig:pccr}
\end{figure*}

\begin{figure*}
    \centering\scalebox{0.5}{
    \includegraphics{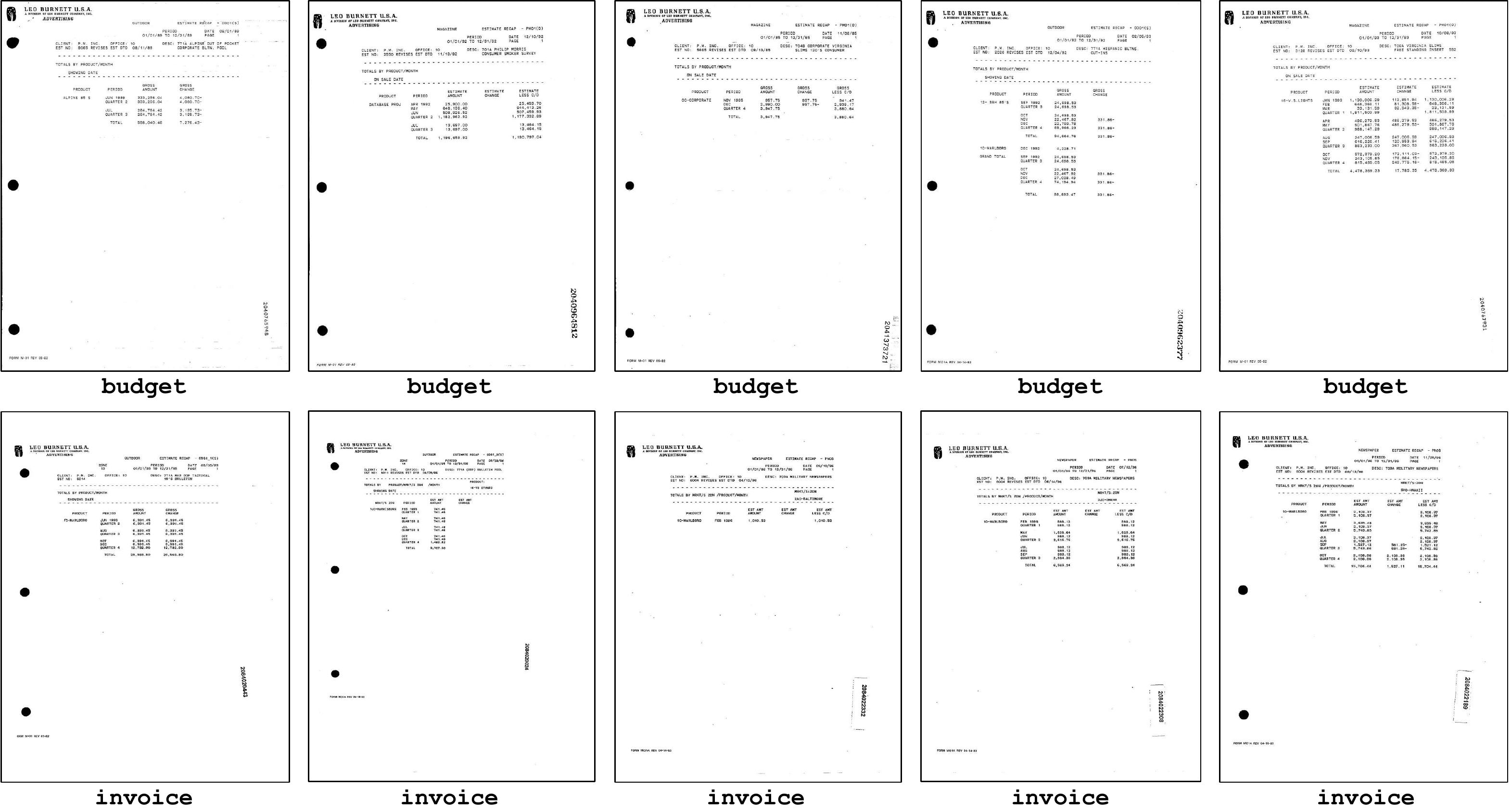}}
    \caption{Examples of advertisement placement report images from the \texttt{budget} (top) and \texttt{invoice} document categories.}
    \label{fig:adv}
\end{figure*}

Figure~\ref{fig:hist-similarities} charts maximum similarity scores between test and train samples for the RVL-CDIP test data.
Figure ~\ref{fig:non-english_samples} lists several non-English samples from RVL-CDIP.
Figures~\ref{fig:more-unknowns}--\ref{fig:more-mixed} show example errors and ambiguous documents.
Figures~\ref{fig:appendix-similarities}--\ref{fig:error_similarities} display test-train pairs with corresponding similarity scores.
Figure~\ref{fig:biosketches} show examples of "Biographical Sketch" documents from the \texttt{resume} category, illustrating the high level of similarity of this particular sub-type; Figure~\ref{fig:specification_templates} shows a similar case for another category.
Figures~\ref{fig:checks}-\ref{fig:adv} show cases where two categories have the same sub-types of documents.


\subsection{Data Availability}\label{sec:data-github}
The data and metadata that we annotated as part of our error analysis (excluding data with sensitive information) is available at: \href{https://github.com/gxlarson/rvlcdip-errors}{\url{github.com/gxlarson/rvlcdip-errors}}.

\end{document}